\documentclass{article}

\PassOptionsToPackage{numbers, compress}{natbib}

\usepackage[final]{neurips_2022}

\usepackage{wrapfig}
\usepackage{comment}
\usepackage{dsfont}
\usepackage[utf8]{inputenc} 
\usepackage[T1]{fontenc}    
\usepackage{hyperref}       
\usepackage{url}            
\usepackage{booktabs}       
\usepackage{amsmath,amsfonts,amssymb}
\usepackage{amsthm}

\usepackage{nicefrac} 
\usepackage{graphicx,subfigure,caption}
\usepackage{microtype}      
\usepackage{xcolor}         
\usepackage{algorithm}
\usepackage{algorithmic}
\usepackage{enumitem}

\title{Learning on Arbitrary Graph Topologies \\ via Predictive Coding}

\author{%
    \textbf{Tommaso Salvatori}$^{1,*}$ \ \ \ \  
    \textbf{Luca Pinchetti}$^{1,*}$\ \ \ \  
    \textbf{Beren Millidge}$^{2}$ \ \ \ \ 
    \textbf{Yuhang Song}$^{1,2,\dag}$ \\
    \textbf{Tianyi Bao}$^{1}$ \ \ \ \  
    \textbf{Rafal Bogacz}$^{2}$\ \ \ \ 
    \textbf{Thomas Lukasiewicz}$^{3,1}$\\
    $^1$\,Department of Computer Science, University of Oxford, UK\\
    $^2$\,MRC Brain Network Dynamics Unit, University of Oxford, UK\\
    $^3$\,Institute of Logic and Computation, TU Wien, Austria  \\
    \texttt{tommaso.salvatori@cs.ox.ac.uk, luca.pinchetti@cs.ox.ac.uk}\\
    \texttt{beren.millidge@ndcn.ox.ac.uk, yuhang.song@some.ox.ac.uk,
    tianyi.bao@cs.ox.ac.uk}\\
    \texttt{rafal.bogacz@ndcn.ox.ac.uk, thomas.lukasiewicz@cs.ox.ac.uk} \\
}

\begin{document}

\maketitle
\renewcommand{\thefootnote}{\fnsymbol{footnote}}
\footnotetext{ \dag\,Corresponding author.}
\footnotetext{ *\,Equal contribution.}

\begin{abstract}
Training with backpropagation (BP) in standard deep learning consists of two main steps: a forward pass that maps a data point to its prediction, and a backward pass that propagates the error of this prediction back through the network. This process is highly effective when the goal is to minimize a specific objective function. However, it does not allow training on networks with cyclic or backward connections. This is an obstacle to reaching brain-like capabilities, as the highly complex heterarchical structure of the neural connections in the neocortex are potentially fundamental for its effectiveness. In this paper, we show how predictive coding (PC), a theory of information processing in the cortex, can be used to perform inference and learning on arbitrary graph topologies. We experimentally show how this formulation, called \emph{PC graphs}, can be used to flexibly perform different tasks with the same network by simply stimulating specific neurons. This enables the model to be queried on stimuli with different structures, such as partial images, images with labels, or images without labels. We conclude by investigating how the topology of the graph influences the final performance, and comparing against simple baselines trained~with~BP.
\end{abstract}

\section{{Introduction}}
\label{sec:intro}

Classical deep learning has achieved remarkable results by training deep neural networks to minimize an objective function. Here, every weight parameter gets updated to minimize this function using reverse differentiation \cite{rumelhart1986learning,linnainmaa1970representation}. However, in the brain, every synaptic connection is independently updated to correct the behaviour of its post-synaptic neuron \cite{hebb49} using local information, and it is unknown whether this process minimizes a global objective function. The brain maintains an internal model of the world, which constantly generates predictions of external stimuli. When the predictions differ from reality, the brain immediately corrects this error (difference between reality and prediction) by updating the strengths of the synaptic connections \cite{srinivasan1982,mumford92,friston2003learning, rao1999predictive}. This theory of information processing, called \emph{predictive coding (PC)}, is highly influential, despite experimental evidence in the cortex being mixed \cite{walsh2020evaluating,kell2018task,millidge2021predictive,bastos2012canonical}, and it is at the centre of a large amount of research in computational neuroscience \cite{friston2017graphical,friston2005theory,spratling2017review,huang2011predictive,friston2008variational}. From the machine learning perspective, PC has  promising properties: it is able to achieve excellent results in classification  \cite{whittington2017approximation} and memorization \cite{salvatori2021associative}, and is able to process information in both a bottom up and a top down direction. This last property is fundamental for the functioning of different brain areas, such as the hippocampus \cite{Barron20,salvatori2021associative}. PC also shares the generalization capabilities of standard deep learning, as it is able to approximate backpropagation (BP) on any neural structure \cite{millidge2020predictive}, and a variation of PC is able to exactly replicate the weight update of BP on any computational graph \cite{Song2020,salvatori2021any}. Moreover, PC only uses local information to update synapses, allowing the network to be fully parallelized, and to train on networks with any~topology.

\begin{figure*}[t]
\medskip 
    \centering
	\includegraphics[width=.8\linewidth]{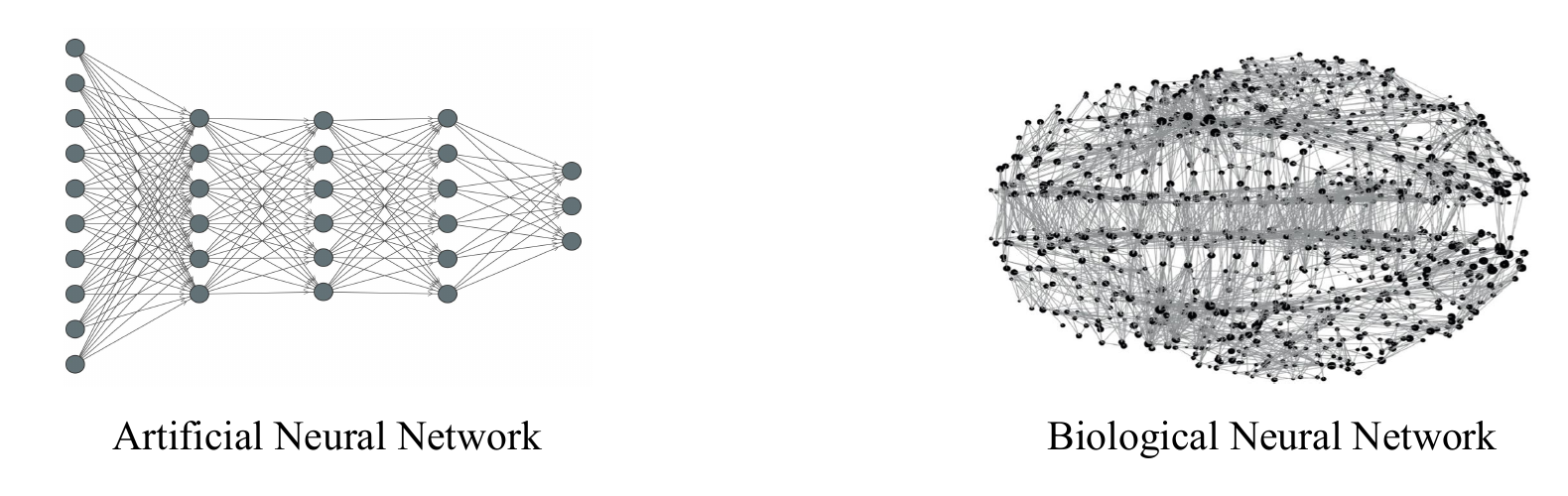}
 \vspace*{-1ex}
  \caption{Difference in topology between an artificial neural network (left), and a sketch of a network of structural connections that link distinct neural elements in a brain (right) \cite{avena18}. }
  \label{fig:brain}
\end{figure*}

Training on networks of any structure is not possible in standard deep learning, where information only flows in one direction via the feedforward pass, and then BP is performed in sequential steps backwards. If a cycle is present inside the computational graph of an artificial neural network (ANN), BP becomes stuck in an infinite loop. More generally, the computational graph of any function $F\colon\mathbb{R}^d \rightarrow \mathbb{R}^k$ is a poset, and hence acyclic. While the problem of training on some specific cyclic structures has been partially addressed using BP through time \cite{lstm} on sequential data, the restriction to hierarchical architectures may present a limitation to reaching brain-like intelligence, since the human brain has an extremely complex and entangled neural structure that is heterarchically organized with small-world connections \cite{avena18}---a topology that is likely highly optimized by evolution. This shape of structural brain networks, shown in Fig.~\ref{fig:brain}, generates a unique communication dynamics that is fundamental for information processing in the brain, as different aspects of network topology imply different communication mechanisms, and hence perform different tasks \cite{avena18}. The heterarchical topology of brain networks has motivated research that aims to develop learning methods on graphs of any topology. A popular example is the \emph{assembly calculus} \cite{Papadimitriou20,dabagia21}, a Hebbian learning method that can perform different operations implicated in cognitive phenomena.


In this work, we address this problem by proposing \emph{PC graphs}, a structure that allows to train on any directed graph using the original (error-driven) framework by Rao and Ballard \cite{rao1999predictive}. We then demonstrate the flexibility of such networks by testing the same network on different tasks, which can be interpreted as conditional expectations on different neurons of the network. Our PC graphs framework enables the model to be queried on stimuli with different structures, such as partial images, images with labels, or images without labels. This is significantly more flexible than the strict input-output structure of standard ANNs, which are limited to scenarios when they are always presented with~data and labels in the same format.

Note that the main goal of this work is not to propose a specific architecture that achieves state-of-the-art (SOTA) results on a particular task, but to present PC graphs as a new flexible and biologically plausible model, which can achieve good results on many tasks simultaneously. 
In this work, we study the simultaneous generation, classification, and associative memory capabilities of PC graphs, highlighting their flexibility and theoretical advantages over standard baselines. Our contributions are briefly summarized as follows:
%
\begin{itemize}[leftmargin=*, itemsep=0.5pt]
\item We present PC graphs, which generalize PC  to arbitrary graph topologies, and show how a single model can be queried in multiple ways to solve different tasks by simply altering the values of specific nodes, without the need for retraining when switching between tasks. Particularly, we define two different techniques, which we call \emph{query by conditioning} and  \emph{query by initialization}. 
\item   We then experimentally  show this in the most general case, i.e., for fully connected PC~graphs. Here, we train different models on MNIST and FashionMNIST, and show how the two queries can be used to perform different generation tasks. Then, we test the model on classification tasks, and explore its capabilities as an associative memory model.
\item   We next investigate how different graph topologies influence the performance of PC graphs on generation tasks, reproducing common network architectures such as feedforward, recurrent, and residual networks as special cases of PC graphs, and investigate how the chosen structure influences the performance on generative tasks. Finally, we also show how PC graphs can be used to derive the popular \emph{assembly calculus}  \cite{Papadimitriou20}.
\end{itemize}

\section{PC Graphs}\label{sec:prem}

\begin{figure*}[t]
\medskip 
    \centering
	\includegraphics[width=\linewidth]{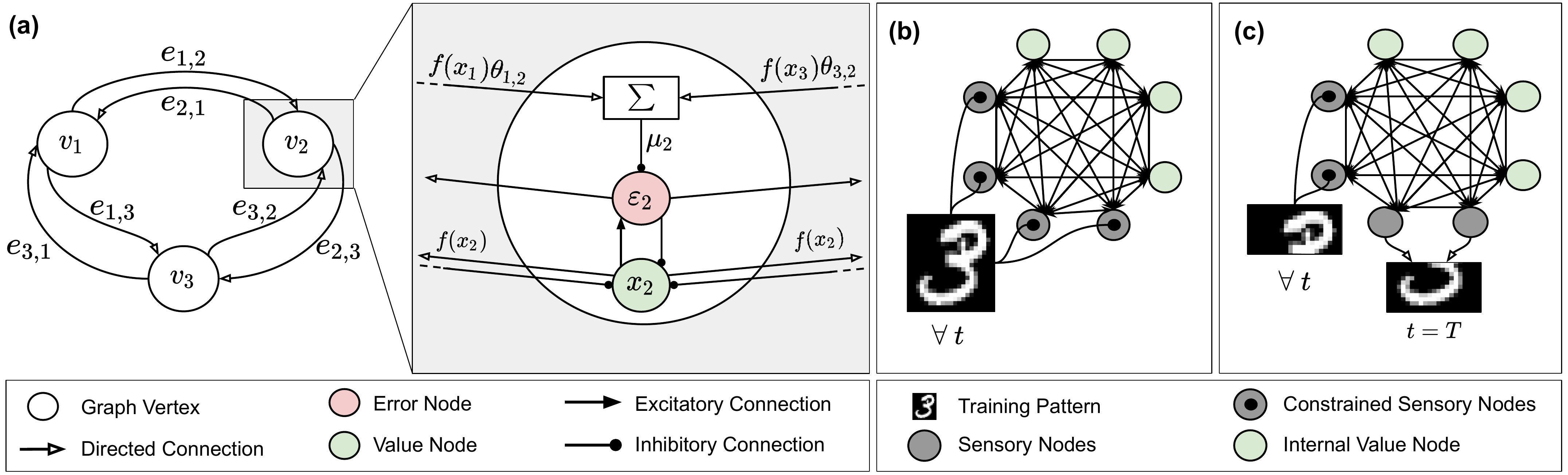}
\caption[short]{(a) An example of a fully connected PC graph with three vertices. Zoomed is the neural implementation of  PC, where learning is made local via the demonstrated inhibitory and excitatory connections. (b) A sketch of the training process, where the value nodes of the sensory vertices are fixed to the pixels of the image. (c) A sketch of query by conditioning, where a fraction of the value nodes is fixed to the top half of an image, and the bottom half is recovered via inference.}
\label{fig:fully}
\end{figure*}

Let $G=(V,E)$ be a directed graph, where $V$ is a set of $n$ vertices $\{1,2,\dots,n\}$, and $E\subseteq V\times V$ is a set of directed edges between them, where every edge $({i,j})\in E$ has a weight parameter~$\theta_{i,j}$. The set of vertices $V$ is partitioned into two subsets, the \emph{sensory} and \emph{internal vertices}. External stimuli are always presented to the network via sensory vertices, which we consider to be the first $d$ vertices of the graph, with $d<n$. The internal vertices, on the other hand, are used to represent the internal structure of the dataset. Each vertex $i$ encodes several quantities. The main quantity is given by  the values of its activity, which change over time, and we refer to it as a \emph{value node} $x_{i,t}$. We call the value nodes of the sensory vertices \emph{sensory nodes}. Additionally, each vertex computes the \emph{prediction} $\mu_{i,t}$ of its activity based on its input from value nodes of other vertices:
\begin{equation}
    \mu_{i,t} = {\textstyle\sum}_{j} \theta_{j,i} f(x_{j,t}),
\end{equation}
where the summation is over all the vertices $j$ connected to $i$ via outgoing edges, and $f$ is a non-linearity. Equivalently, it is possible to consider the summation on every $j$, and have $\theta_{i,j} = 0$ if $({i,j}) \not\in E$.  The error of every vertex at every time step $t$ is then given by the difference between its value node and its prediction, i.e., $\varepsilon_{i,t} = x_{i,t} - \mu_{i,t}$. 
This local definition of error, which lies not only in the output, but in every vertex of the network, is what allows PC graphs to learn using only local information. The value nodes $x_{i,t}$ and the weight parameters $\theta_{i,j}$ are updated to minimize the following energy function defined locally on every vertex:
\begin{equation}
    \mathcal{E}_{t} = { \mbox{$\frac{1}{2}$} {\textstyle\sum}_{i}  ( \varepsilon_{i,t} ) ^2}.
\label{eq:energy}
\end{equation}%
\noindent A fully connected PC graph with $3$ vertices is sketched in Fig.~\ref{fig:fully}a, along with the operations that describe the dynamics of the information flow, showing also how every operation can be represented via inhibitory and excitatory connections.

\textbf{Learning:} When presented with a training point $\bar s$ taken from a training set, the value nodes of the sensory vertices are fixed to be equal to the entries of $\bar s$ for the whole duration of the training process, i.e., for every $t$. A sketch of this is shown in Fig.~\ref{fig:fully}b. Then, the total energy of Eq.~\eqref{eq:energy} is minimized in two phases: \emph{inference} and \emph{weight update}. During the inference phase, the weights are fixed, and the value nodes are continuously updated via gradient descent for $T$ iterations, where $T$ is a hyperparameter of the model. The update rule is the following (\textit{inference}):
\begin{align}
\Delta{x}_{i,t} = - \gamma \cdot \partial \mathcal{E}_t/\partial x_{i,t} = \gamma\cdot ( -\varepsilon_{i,t} + f' ( x_{i,t} ) {\textstyle\sum}_{k=1}^{n} \varepsilon_{k,t} \theta_{k,i}),
\label{eq:x_update}
\end{align}
where $\gamma$ is the learning rate of the value nodes. This process of iteratively updating the value nodes distributes the output error throughout the PC graph. When the inference phase is completed, the value nodes get fixed, and a single weight update is performed as follows (\textit{weight update}): 
\begin{align}\label{eq:theta_update}
\Delta \theta_{i,j}  &= -\alpha\cdot {\partial \mathcal{E}_T}/{\partial \theta_{i,j}} =  \alpha\cdot \varepsilon^l_{i,T} f ( x_{j,T}),
\end{align}
where $\alpha$ is the learning rate of the weight update. We now describe two different ways to query the internal representation of a trained model, where the values of some sensory vertices are undefined, and have to be predicted. In both cases, the weight parameters $\theta_{i,j}$ are now fixed, and the total energy $E$ is continuously minimized using gradient descent on the re-initialized value nodes via Eq.~\eqref{eq:x_update}.

\textbf{Query by conditioning:} While each value node is randomly re-initialized, the value nodes of specific vertices are \emph{fixed} to some desired value, and hence not allowed to change during the energy minimization process. The unconstrained sensory vertices will then converge to the minimum of the energy given the fixed vertices, thus computing the conditional expectation of the latent vertices given the observed stimulus. Formally, let $I = \{i_1,\dots, i_q\} \subset \{1,2,\ldots,n\}$ be a strict subset of vertices. Assume now that we know that a subset of the value nodes corresponding to the vertices $I$ is equal to a stimulus $\bar q \in \mathbb{R}^q$. Then, running inference until convergence allows to estimate the conditional expectation
\begin{align}
    E(\bar x_T \mid \forall t\colon (x_{i_1,t},\dots,x_{i_q,t}) = \bar q ),
\end{align}
where $\bar x_T$ is the vector of value nodes at convergence. Examples of tasks performed this way are (i)~classification, where internal nodes are fixed to the pixels of an image, and the sensory nodes are fixed to a $1$-hot vector with the labels, (ii) generation, where the single value node encoding the class information is fixed, and the value nodes of the sensory nodes converge to an image of that class, and (iii) reconstruction, such as image completion, where a fraction of the sensory nodes are fixed to the available pixels of an image, and the remaining ones converge to a reasonable completion of it. A~sketch of this process is shown in Fig.~\ref{fig:fully}c.

\textbf{Query by initialization:} Again, every value node is randomly initialized, but the value nodes of specific nodes are \emph{initialized} (for $t\,{=}\,0$ only), but not fixed (for all $t$), to some desired value. This differs from the previous query, as here every value node is unconstrained, and hence free to change during inference. The sensory vertices will then converge to the minimum found by gradient descent, when provided with that specific initialization. Again, let $I = \{i_1,\dots, i_q\} \subset$ 
$\{1,2,\ldots,n\}$ be a strict subset of vertices, and  assume that we have an initial stimulus $\bar q \in \mathbb{R}^q$. Then, we can estimate the conditional expectation
\begin{align}
    E(\bar x_T \mid (x_{i_1,0},\dots,x_{i_q,0}) = \bar q)\,.
\end{align}
Examples of tasks performed this way are (i) denoising, such as image denoising, where the sensory neurons are initialized with a noisy version of an image, which is cleared during the energy minimization process, and (ii) reconstruction, such as 
image completion, where the fraction of missing~pixels is now not known a priori.



\section{Proof-of-concept: Experiments on Fully Connected PC Graphs}

\begin{figure*}[t]
\medskip 
    \centering
	\includegraphics[width=1.0\textwidth]{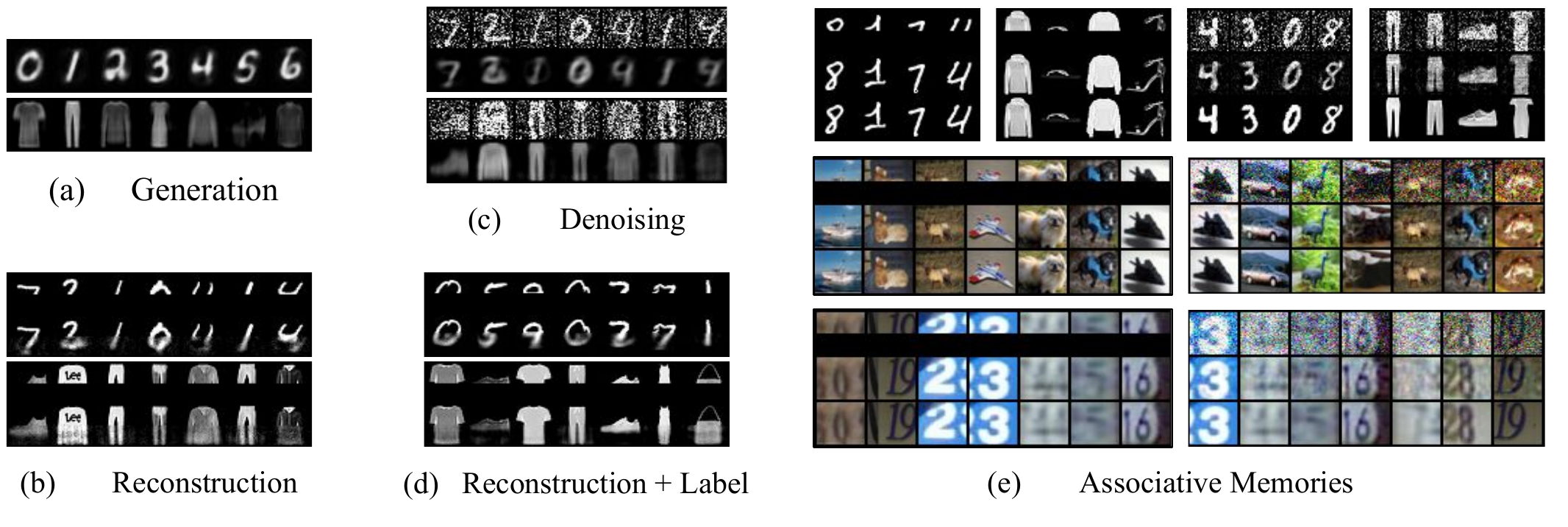}
	\vspace*{-3ex}
\caption[short]{ Generation experiments using the first $6$ classes of the MNIST and FashionMNIST datasets from the labels $\{$0, 1, 2, 3, 4, 5, 6$\}$ and $\{$t-shirt, trouser, pullover, dress, coat, sandal, shirt$\}$, respectively; (b) reconstruction of incomplete images using \emph{query by conditioning}, when only the top half is available; (c) reconstruction of corrupted images using \emph{query by initialization}; (d)~reconstruction of incomplete images using \emph{query by conditioning} when also providing the correct label of the test image; and (e) associative memory experiments when presented with half of a training image (left) or a corrupted version (right) that it has already seen and memorized; from top to bottom row: image provided to the network, retrieved image, and original image.}
\label{fig:Gen}
\end{figure*}

In this section, we perform experiments on a fully connected PC graph $G=(V,E)$, i.e., where $E=V\times V$. Such PC graphs are fully general and encode no implicit priors on the structure of the dataset. It is possible to obtain any possible graph topology by simply pruning specific weights of $G$.

Given a dataset of $m$ datapoints $\mathcal D = \{ \bar s_i \}_{i<m}$, with  $\bar s_i \in \mathbb{R}^d$, we train the PC graph as described in Section 2: The first $d$ neurons are fixed  to the entries of a training point, and the energy function $\mathcal{E}_t$ is minimized via inference and weight updates, via  Eqs.~\eqref{eq:x_update} and \eqref{eq:theta_update}. When the training is complete, we show the different tasks that can be performed, without the need of retraining the model. We use MNIST  and FashionMNIST  \cite{fashionmnist}, fixing the first $d$ nodes to the data point, and show how to perform the tasks of generation, denoising, 
reconstruction (without and with labels),  and classification by querying the PC graph as described in Section~\ref{sec:prem}.

\textbf{Setup:} For every dataset, we have trained $3$ models: one for generation and classification tasks, one for denoising and reconstructions, and one for associative memories. The first two models consist of a fully connected graph with $2000$ vertices, trained with $794$ sensory vertices for classification and generation tasks ($784$ pixels plus  a 1-hot vector for the $10$ labels), and $784$ sensory vertices for reconstruction and denoising. Further details about other hyperparameters are given in the supplementary material. 

\textbf{Generation:} To check the generation capabilities of a trained PC graph, we queried the model by conditioning on the labels: Here, the value nodes dedicated to the $10$ labels were fixed to each 1-hot value, and the energy of the model (Eq.~\eqref{eq:energy}) was minimized using Eq.~\eqref{eq:x_update} until convergence. The generated images are then taken to be the value nodes of the unconstrained sensory nodes, which were originally fixed to the pixels of the images during training. An example of the images generated for each label is given in Fig.~\ref{fig:Gen}a.

\textbf{Reconstruction:} We provide the PC graph with half of a test image, and ask it to reconstruct the second half. This can be done using both queries: when querying by conditioning, half of the pixels of a test image are fixed to the corresponding sensory nodes; when querying by initialization, the value nodes are simply initialized to the same values. At convergence, we consider the value nodes of the unconstrained nodes, which should reconstruct the missing part of the image based on the information learned during training. The results are given in Fig.~\ref{fig:Gen}b. We have also replicated the same experiment using a network trained with the labels, and provided the label during the reconstruction. This computes the distribution of the missing pixels knowing the available ones \emph{and} the label. The results in this case are visibly better and are given in Fig.~\ref{fig:Gen}d.

\textbf{Denoising:} We provide the PC graph with  a corrupted image, obtained by adding zero-mean Gaussian noise with variance $0.5$. This is done by querying by initialization: before running inference, the value nodes of the sensory nodes are initialized to be equal to the pixels of the corrupted image. At convergence, we consider the value nodes of the unconstrained nodes, which should reconstruct the original image. The results are given in Fig.~\ref{fig:Gen}c.

\textbf{Results:} As stated above, we picked a fully connected PC graph due to its generality, and not to obtain the best performance. However, the results show that this framework is able to learn an internal representation of a dataset, and that it can be queried to solve multiple tasks with a reasonable accuracy. The PC graph was in fact able to always generate the correct digit, and almost always able to generate the correct clothing item in generation tasks, and always able to provide a noisy but reasonable reconstruction of incomplete test points. The same happened with denoising experiments, as a cleaner (plausible) image was always produced. In Section 4, we show how to improve all these performances by using different PC graph topologies.

\begin{table*}[t]
    \centering
	\caption{Test accuracy of different models on MNIST, FashionMNIST, and SVHN.}
	
	\smallskip
	\begin{tabular}{lcccc}
		\toprule
		 Model & Ours & Hopfield Nets & Boltzmann Machine & Almeida Pineda   \\
		\toprule
		MNIST  & 91.76 $\pm$ 0.02 \%  & 65.23 $\pm$ 2.21 \% & 79.23 $\pm$ 0.15 \% & 76.36 $\pm$ 0.14 \%  \\
		FashionMNIST  & 83.72 $\pm$ 0.33 \% & 51.74 $\pm$ 3.94 \% & 61.31 $\pm$ 0.17 \% & 69.63 $\pm$ 1.64 \% \\
		SVHN  & 84.51 $\pm$ 0.11 \%   & 48.92 $\pm$ 3.11 \% & 55.74 $\pm$ 1.23 \% & 59.14 $\pm$ 2.64 \% \\
		\bottomrule
	\end{tabular}
	\label{tab:class}
\end{table*}

\textbf{Classification:} We consider the same PC graph trained for the generation experiments. To check its generalization capabilities, we query by conditioning the pixels of every test image to the first $784$ sensory nodes, and run inference to reconstruct the $1$-hot label vector. We do not expect to obtain results directly comparable with standard multilayer perceptrons for two reasons: firstly, the model does not contain any implicit hierarchy, which empirically appears crucial to obtaining good classification results. Secondly, the PC graph is also simultaneously learning to generate the pixels, which are much more numerous than labels.  However, to check whether the obtained results were acceptable, we tested against different learning algorithms that train on similar or equivalent fully connected architectures, such as Hopfield networks, unconstrained Boltzmann machines, and a local variation of BP introduced in the late $'80$, called Almeida-Pineda, named after the two scientists who independently invented it \cite{almeida90,pineda87}. As for Hopfield networks, we 
used the implementation provided in \cite{belyaev20}. The results,  given in Table~\ref{tab:class}, show that our model outperforms every other learning algorithm that can be trained on fully connected architectures. Despite this, the results also show that the obtained test accuracy is not nearly comparable to the results obtained by multilayer perceptrons, as they are only slightly better than a linear classifier (obtaining $88\%$ accuracy on MNIST). However, this is not due to the learning rule of PC, which is 
well-known to be able to reach a competitive performance when provided with a hierarchical  multilayer structure \cite{whittington2017approximation}. For the SVHN \cite{svhn}
experiment, we used models with $5000$~vertices.

\begin{figure*}[t]
\medskip 
    \centering
	\includegraphics[width=\textwidth]{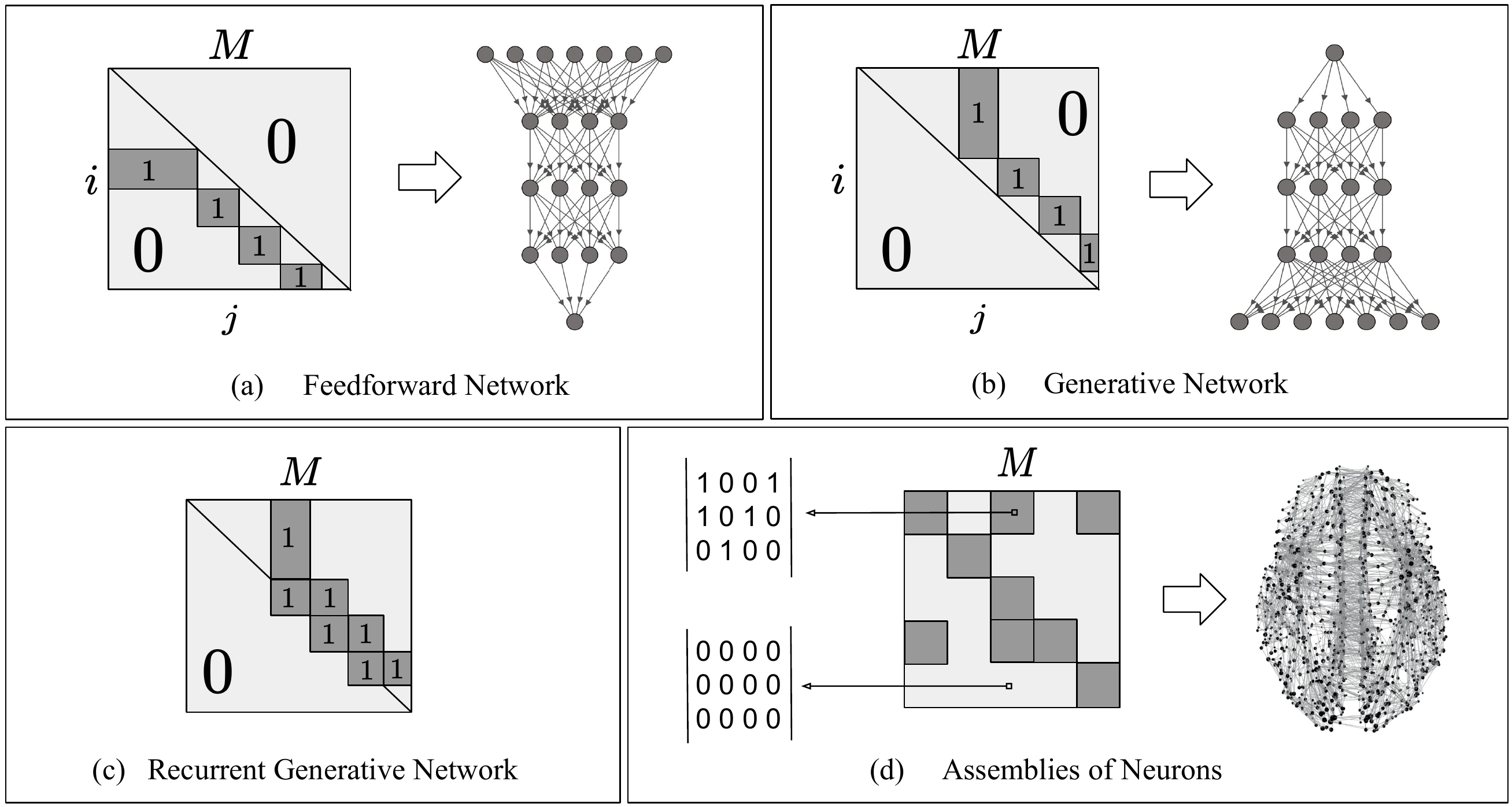}
\caption{Examples of PC graphs that can be built by masking a part of the weights of a fully connected PC graph. (a) Masking required to build a standard multilayer architecture, such as the one in \cite{whittington2017approximation}. (b) Masking required to build a multilayer architecture, where the weights go in the opposite direction. Here, the sensory nodes are at the end of the hierarchical structure. This model is equivalent to the generative networks in \cite{salvatori2021associative}. (c) Examples of masking needed to implement popular architectures with lateral connections, similar to the model  in \cite{ororbia2020}. (d) This is the model  in \cite{Papadimitriou20}, which consists of a set of Erdõs–Renyi graphs that simulate brain regions (dark squares on the diagonal) and connnections between them (dark squares off the diagonal).}
\label{fig:models}
\end{figure*}

\textbf{Associative memory:} We now test whether PC graphs are able to memorize training images and retrieve them given a corrupted or incomplete version of it. Particularly, we show that a fully connected PC graph is able to store complex data points, such as colored images, and retrieve them via running inference. To do that, we trained a novel fully connected PC graph on $100$ data points of the MNIST, FashionMNIST, CIFAR10, and SVHN datasets. We have used a model with $1000$ vertices for MNIST and FashionMNIST, and $3500$ for SVHN and CIFAR10, and asked it to retrieve the original memories by presenting it either only half of the original pixels, or a corrupted version with Gaussian noise variance $0.2$. This task is similar to image reconstruction and denoising, with the non-trivial difference that here we only use already seen data points, and hence no generalization is involved. The results of these experiments are given in Fig.~\ref{fig:Gen}e, and show that our method is able to successfully store and retrieve~data~points via energy minimization. More details about the capacity of fully connected PC graphs are given in the supplementary material.

\section{Extension to Different PC Graph Topologies}

As well-known in deep learning, the performance of the trained model strongly depends on its architecture: the number of vertices, layers, and their intrinsic structure.  In Section 3, we studied the general architecture of  fully connected PC graphs. Here, we show how  to reduce a fully connected PC graph to lighter and even more powerful PC graphs. Particularly, we show how to generate different neural architectures by simply pruning specific edges of a fully connected PC graph $G=(V,E)$. In this case, the pruning is performed by applying a sparse mask $M$. However, there are multiple~equivalent ways of implementing it. 
Consider now the weight matrix $\bar \theta \in \mathbb{R}^{n \times n}$, where every entry $\theta_{i,j}$ represents the weight parameter connecting vertex $i$ to vertex $j$. To generate a neural architecture that consists of a subset of the original connections, it suffices to \emph{mask} the matrix $\bar \theta$ via entry-wise multiplication with a binary matrix $M$, where $M_{i,j} = 1$ if the edge $({i,j})$ exists in $E$, and $M_{i,j} = 0$ otherwise. This allows the creation of hierarchical discriminative architectures such as a PC equivalent of the multilayer perceptron (MLP) in Fig.~\ref{fig:models}a, or hierarchical generative networks in Fig.~\ref{fig:models}b,\,c. More generally, it creates a framework to generate and study architectures with any topology, such as small-world networks inspired by brain regions \cite{telesford2011ubiquity}, as shown in Fig.~\ref{fig:models}d. Guidance on which topology should be used depends on the tasks and dataset, and it is hence hard to propose a general theory (as it is with BP). In what follows, however, we provide multiple examples.

\textbf{Experiments:} Here, we study how the network topology influences the final performance, performing the same experiments shown on the fully connected PC graph. We expect the generated images to be visibly better due to the enforced hierarchical structure of the PC~graph.


\textbf{Setup:} We  trained generative PC graphs, recurrent generative PC graphs, assemblies of neurons PC graphs, and standard BP autoencoders with different numbers of hidden layers and hidden dimension, and report the best results. For the generation results, we  used the same setup, but added an input layer with $10$ vertices, whose value nodes during training were initialized with the 1-hot label vector. We performed a search across learning rates $\gamma$ and $\alpha$, and on the number of iterations per batch $T$. More details are given in the supplementary material, as well as a long discussion on how different parameters influence the final performance of the architecture.

\begin{figure*}[t]
\medskip 
    \centering
	\includegraphics[width=1.0\textwidth]{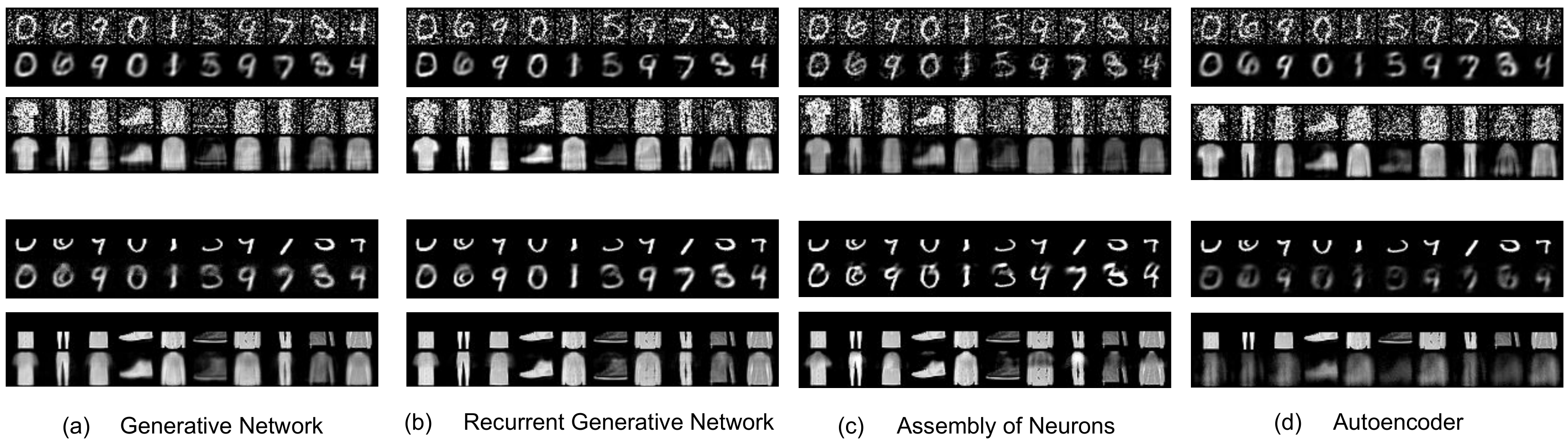}
	\vspace*{-3ex}
\caption[short]{\emph{Query by initialization} (top) and \emph{query by conditioning} (bottom) on three different PC graph architectures and different datasets. Particularly, we tested these PC graphs against ANN autoencoders trained with BP (d), which perform comparably to the PC graphs on denoising tasks, but less well on image reconstruction.
}
\label{fig:any}
\end{figure*}

\textbf{Results:}  The results are given in Fig.~\ref{fig:any}a and b. As expected, the hierarchical structure of the considered PC graphs improves over the fully connected PC graph, despite being comparable in the number of parameters. Compared against autoencoders (Fig.~\ref{fig:any}c), the standard ANN baseline trained~with BP, the PC graph results are similar in image denoising, and better in image reconstruction. FID scores on denoising tasks for different levels of noise are given in Table \ref{tab:fid}. 

\begin{figure*}[t]
\medskip 
    \centering
	\includegraphics[width=1.0\textwidth]{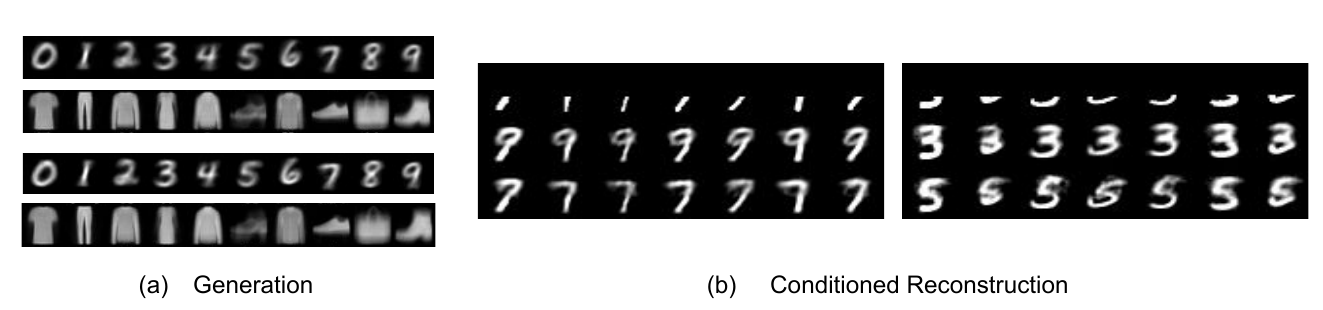}
  \caption{Left: Generated images given the labels using feedforward (top) and recurrent (bottom) PC graphs. Right: conditional inference on the labels.}
  \label{fig:any_gen}
\end{figure*}

\section{Conditioning on Labels}

Assume that we need to reconstruct a test image from an incomplete version of it, with the further assumption that that this time we are also provided with the label of the corrupted image. It would be useful to be able use this extra information to obtain a better reconstruction. In PC graphs, this is straightforward: it suffices to simultaneously fix the value nodes representing the labels to the 1-hot vector of the provided label, and the sensory nodes to the pixels of the corrupted image. This method can be applied when it is difficult to infer to which class an incomplete image belongs, and providing the label during the reconstruction allows the preferred label to influence the reconstruction. Hence, we perform the following task: we provide images of digits that look similar when incomplete, and ask the model to reconstruct the missing half when giving the label information, i.e., use the additional label information to correctly resolve the inherent ambiguity in the reconstruction task.

\textbf{Experiments:} We used the same PC graphs from above for generation tasks. We provided the PC graph the bottom $33\%$ of random images representing $7$s or $9$s. Note that it is hard to distinguish between these two numbers when only this small portion of the image is available. Then, we generated the missing $67\%$ of the pixels by first giving $7$ as a label, and then giving $9$. We have repeated the same task using $3$s and $5$s. The results, available in Fig~\ref{fig:any_gen}b, show that our model is able to perform conditional inference, as the reconstructed digits always agree with the provided labels.

\begin{wrapfigure}{r}{0.36\textwidth}
\vspace*{-9ex}
  \caption{FID Score on MNIST on images corrupted with Gaussian noise of different variance.}
  \centering
	\begin{tabular}{lcc}
		\toprule
		 Method & PC & Autoencoder    \\
		\toprule
		0.2  & 25.61   & 43.93  \\
		0.5  & 44.53  & 53.79  \\
		0.7  & 51.38   & 57.56 \\
		\bottomrule
	\end{tabular}
\label{tab:fid}
\end{wrapfigure}

\section{Assembly of Neurons}

Recently, a model made by assemblies of neurons that are sparsely connected with each other has been proposed to emulate brain regions \cite{Papadimitriou20}. This model consists of $m$ ordered clusters of neurons $(C_1,\dots,C_m)$, and any two ordered neurons of the same cluster are connected by a synapse with probability $p$, creating an Erdõs–Renyi graph $G_{m,p}$. Depending on the desired task, two clusters can be connected via sparse connections following the same rule: if cluster $C_a$ is connected to cluster $C_b$, then, given a vertex $v_i \in C_a$ and a  vertex $v_j \in C_b$, there exists a synaptic connection connecting $v_i$ to $v_j$ with probability $p$. Note that this structure is highly general, and allows to build networks such as the one represented in Fig.~\ref{fig:brain}b. To conclude, at each time step, only the $k$ neurons of every cluster with the highest neural activity fire. In the original work, the authors propose a Hebbian-like learning algorithm, however, we show that it can also be trained using PC graphs. A~graphical representation on how to encode as a PC graph a network made by assemblies of neurons is given in Fig.~\ref{fig:models}d. In this case, each dark block on the diagonal represents connections between neurons of the same region. Unlike the other networks in the same figure, these are sparse matrices where every entry is either zero, or one with probability $p$. As in the brain, not every region is connected with the other, and whether two regions are directly connected has to be decided a priori when designing the architecture. Again, two neurons between connected regions are directly connected with probability $p$. In Fig.~\ref{fig:models}d,  dark blocks off the diagonal represent the presence of directed connections between two regions $C_a$ and $C_b$. If situated below the diagonal, the connections go from $C_a$ to $C_b$, with $a<b$; if situated above the diagonal, they go from $C_b$ to~$C_a$.

\textbf{Experiments:} We replicated this structure, using $4$ clusters with $3000$ vertices each, connected in a feedforward way: the first cluster is connected with the second, which is connected with the third, and so on. As sparsity and top-k constants, we  used $p=0.1$ and $k=0.2$, and performed the same generative experiments. The results are given in Fig.~\ref{fig:any}c. While the results look cleaner than the other methods, note that they are specific to MNIST and FashionMNIST, as the top-k activation on the last cluster well cleans the noise surrounding the reconstructions.


\vspace*{-1ex}
\section{Related Work}
\vspace*{-1ex}

Our work shares similarities and the final goal with a whole field of research that aims to improve current neural networks by using techniques from  computational neuroscience. In fact, the biological implausibility and limitations of BP highlighted in \cite{Lillicrap20,whittington2019theories} have fueled research in finding a new learning algorithm to train ANNs, with the most promising candidates being energy-based models such as \emph{equilibrium propagation} \cite{scellier17,scellier2018generalization}. Other interesting energy-based methods are Boltzmann machines \cite{rbm,salakhutdinov2009deep,hinton2009deep}, and Hopfield networks \cite{Hopfield82,Hopfield84}. These differ from PC, as they do not encode the concept of \emph{error}, but learn in a pure Hebbian fashion. Furthermore, they have undirected synaptic connections, and make predictions by minimizing a physical system initialized with a specific input. This is different from PC, that has directed synaptic connections and is tested by fixing specific nodes to an input, while letting the latent ones converge. The PC literature ranges from psychology to neuroscience and machine learning. Particularly, it offers a single mechanism that accounts for diverse perceptual phenomena observed in the brain, examples of which are end-stopping~\cite{rao1999predictive}, repetition-suppression~\cite{auksztulewicz2016repetition}, illusory motions~\cite{lotter2016deep,watanabe2018illusory}, bistable perception~\cite{hohwy2008predictive,weilnhammer2017predictive}, and even attentional modulation of neural activity~\cite{feldman2010attention,kanai2015cerebral}, and it has even been used to describe the retrieval and storage of memories in the human  memory system \cite{Barron20}.

Although inspired by neuroscience models of the cortex, the computational model introduced by Rao and Ballard \cite{rao1999predictive} still presents some implausibilities, with the main one being the presence of symmetric connections. An implementation of PC with no symmetric connections that is able to successfully learn image classification tasks has been presented in \cite{millidge2020relaxing}, and in the \emph{neural generative coding} models, used for continual learning, generative models, and reinforcement learning \cite{ororbia2019biologically,ororbia2022active}.

\vspace*{-1ex}
\section{Discussion}
\vspace*{-1ex}
In this work, we have shown that PC is able to perform machine learning tasks on graphs of any topology, called PC graphs. Particularly, we have highlighted two main differences between our framework and standard deep learning: flexibility in structure and query. On the one hand, a flexible structure allows for learning on any graph topology, hence including both classical deep learning models, and small-world networks that resemble sparse brain regions. On the other hand, flexible querying allows the model to be trained and tested on data points that carry different kinds of information: supervised signals, unsupervised, and incomplete. On a much broader level, this work strengthens the connection between the machine learning and the neuroscience communities, as it underlines the importance of PC in both areas, both as a highly plausible algorithm to train brain-inspired architectures, and as an approach to solve corresponding problems 
in machine intelligence. 

The research of this paper (and current PC literature in general)  is also of great importance from another perspective: training modern neural networks with BP has become computationally extremely expensive, making modern technologies inaccessible. Biological neural networks, on the other hand, do not have these drawbacks thanks to their biological hardware. Recent breakthroughs in the development of neuromorphic and analog computing, such as the finding of the \emph{missing memristor} \cite{strukov2008missing}, could allow the training of deep neural models using only a tiny fraction of energy and time that modern GPUs need. To do this, however, we need to train neural networks end-to-end on the same chip, something that is not possible using BP (or BP through time), due to the need of a control signal that passes information between different layers. The energy formulation of neuroscience-inspired models allows to overcome this limitation, making them perfect candidates to train deep neural models end-to-end on the same chip \cite{kendall2020training}. This strongly motivates research in PC and other neuroscience-inspired algorithm, with a potentially huge long-term impact.

\section*{Acknowledgments}
This work was supported by the Alan Turing Institute under the EPSRC grant EP/N510129/1, 
by the AXA Research Fund, the EPSRC grant EP/R013667/1, the MRC grant MC\textunderscore UU\textunderscore 00003/1, the BBSRC grant BB/S006338/1, and by the EU TAILOR grant. We also acknowledge the use of the EPSRC-funded Tier 2 facility
JADE (EP/P020275/1) and GPU computing support by Scan Computers International Ltd.
Yuhang Song was supported by the China Scholarship Council under the State Scholarship Fund and by a J.P.~Morgan AI Research Fellowship.

\bibliography{neurips_2022}

\begin{thebibliography}{10}

\bibitem{rumelhart1986learning}
D.~E. Rumelhart, G.~E. Hinton, and R.~J. Williams, ``Learning representations
  by back-propagating errors,'' {\em Nature}, vol.~323, no.~6088, pp.~533--536,
  1986.

\bibitem{linnainmaa1970representation}
S.~Linnainmaa, ``The representation of the cumulative rounding error of an
  algorithm as a {T}aylor expansion of the local rounding errors,'' {\em
  Master's Thesis (in Finnish), Univ. Helsinki}, pp.~6--7, 1970.

\bibitem{hebb49}
D.~Hebb, {\em The Organization of Behavior}.
\newblock Wiley, New York, 1949.

\bibitem{srinivasan1982}
M.~V. Srinivasan, S.~B. Laughlin, and A.~Dubs, ``Predictive coding: {A} fresh
  view of inhibition in the retina,'' {\em Proceedings of the Royal Society of
  London. Series B. Biological Sciences}, vol.~216, no.~1205, pp.~427--459,
  1982.

\bibitem{mumford92}
D.~Mumford, ``On the computational architecture of the neocortex,'' {\em
  Biological Cybernetics}, vol.~66, no.~3, pp.~241--251, 1992.

\bibitem{friston2003learning}
K.~Friston, ``Learning and inference in the brain,'' {\em Neural Networks},
  vol.~16, no.~9, pp.~1325--1352, 2003.

\bibitem{rao1999predictive}
R.~P. Rao and D.~H. Ballard, ``Predictive coding in the visual cortex: {A}
  functional interpretation of some extra-classical receptive-field effects,''
  {\em Nature Neuroscience}, vol.~2, no.~1, pp.~79--87, 1999.

\bibitem{walsh2020evaluating}
K.~S. Walsh, D.~P. McGovern, A.~Clark, and R.~G. O'Connell, ``Evaluating the
  neurophysiological evidence for predictive processing as a model of
  perception,'' {\em Annals of the New York Academy of Sciences}, vol.~1464,
  no.~1, p.~242, 2020.

\bibitem{kell2018task}
A.~J. Kell, D.~L. Yamins, E.~N. Shook, S.~V. Norman-Haignere, and J.~H.
  McDermott, ``A task-optimized neural network replicates human auditory
  behavior, predicts brain responses, and reveals a cortical processing
  hierarchy,'' {\em Neuron}, vol.~98, 2018.

\bibitem{millidge2021predictive}
B.~Millidge, A.~Seth, and C.~L. Buckley, ``Predictive coding: A theoretical and
  experimental review,'' {\em arXiv:2107.12979}, 2021.

\bibitem{bastos2012canonical}
A.~M. Bastos, W.~M. Usrey, R.~A. Adams, G.~R. Mangun, P.~Fries, and K.~J.
  Friston, ``Canonical microcircuits for predictive coding,'' {\em Neuron},
  vol.~76, no.~4, pp.~695--711, 2012.

\bibitem{friston2017graphical}
K.~J. Friston, T.~Parr, and B.~de~Vries, ``The graphical brain: Belief
  propagation and active inference,'' {\em Network Neuroscience}, vol.~1,
  no.~4, pp.~381--414, 2017.

\bibitem{friston2005theory}
K.~Friston, ``A theory of cortical responses,'' {\em Philosophical Transactions
  of the Royal Society B: Biological Sciences}, vol.~360, no.~1456, 2005.

\bibitem{spratling2017review}
M.~W. Spratling, ``A review of predictive coding algorithms,'' {\em Brain and
  Cognition}, vol.~112, pp.~92--97, 2017.

\bibitem{huang2011predictive}
Y.~Huang and R.~P. Rao, ``Predictive coding,'' {\em Wiley Interdisciplinary
  Reviews: Cognitive Science}, vol.~2, no.~5, pp.~580--593, 2011.

\bibitem{friston2008variational}
K.~J. Friston, N.~Trujillo-Barreto, and J.~Daunizeau, ``{DEM: A} variational
  treatment of dynamic systems,'' {\em Neuroimage}, vol.~41, no.~3,
  pp.~849--885, 2008.

\bibitem{whittington2017approximation}
J.~C. Whittington and R.~Bogacz, ``An approximation of the error
  backpropagation algorithm in a predictive coding network with local {H}ebbian
  synaptic plasticity,'' {\em Neural Computation}, vol.~29, no.~5, 2017.

\bibitem{salvatori2021associative}
T.~Salvatori, Y.~Song, Y.~Hong, L.~Sha, S.~Frieder, Z.~Xu, R.~Bogacz, and
  T.~Lukasiewicz, ``Associative memories via predictive coding,'' in {\em
  Advances in Neural Information Processing Systems}, vol.~34, 2021.

\bibitem{Barron20}
H.~C. Barron, R.~Auksztulewicz, and K.~Friston, ``Prediction and memory: A
  predictive coding account,'' {\em Progress in Neurobiology}, vol.~192,
  p.~101821, 2020.

\bibitem{millidge2020predictive}
B.~Millidge, A.~Tschantz, and C.~L. Buckley, ``Predictive coding approximates
  backprop along arbitrary computation graphs,'' {\em arXiv:2006.04182}, 2020.

\bibitem{Song2020}
Y.~Song, T.~Lukasiewicz, Z.~Xu, and R.~Bogacz, ``Can the brain do
  backpropagation? --- {E}xact implementation of backpropagation in predictive
  coding networks,'' in {\em Advances in Neural Information Processing
  Systems}, vol.~33, 2020.

\bibitem{salvatori2021any}
T.~Salvatori, Y.~Song, T.~Lukasiewicz, R.~Bogacz, and Z.~Xu, ``Reverse
  differentiation via predictive coding,'' in {\em Proc.~AAAI}, 2022.

\bibitem{avena18}
A.~Avena-Koenigsberger, B.~Misic, and O.~Sporns, ``Communication dynamics in
  complex brain networks,'' {\em Nature Reviews Neuroscience}, vol.~19, no.~1,
  pp.~17--33, 2018.

\bibitem{lstm}
S.~Hochreiter and J.~Schmidhuber, ``Long short-term memory,'' {\em Neural
  Computation}, vol.~9, no.~8, 1997.

\bibitem{Papadimitriou20}
C.~H. Papadimitriou, S.~S. Vempala, D.~Mitropolsky, M.~Collins, and W.~Maass,
  ``Brain computation by assemblies of neurons,'' {\em Proceedings of the
  National Academy of Sciences}, 2020.

\bibitem{dabagia21}
M.~Dabagia, C.~H. Papadimitriou, and S.~S. Vempala, ``Assemblies of neurons can
  learn to classify well-separated distributions,'' {\em arXiv:2110.03171},
  2021.

\bibitem{fashionmnist}
H.~Xiao, K.~Rasul, and R.~Vollgraf, ``{Fashion-MNIST: A} novel image dataset
  for benchmarking machine learning algorithms,'' {\em arXiv:1708.07747}, 2017.

\bibitem{almeida90}
L.~B. Almeida, ``A learning rule for asynchronous perceptrons with feedback in
  a combinatorial environment,'' pp.~102--111, 1990.

\bibitem{pineda87}
F.~J. Pineda, ``Generalization of back-propagation to recurrent neural
  networks,'' {\em Physical Review Letters}, vol.~59, 1987.

\bibitem{belyaev20}
M.~A. Belyaev and A.~A. Velichko, ``Classification of handwritten digits using
  the {H}opfield network,'' in {\em IOP Conference Series: Materials Science
  and Engineering}, vol.~862, IOP Publishing, 2020.

\bibitem{svhn}
Y.~Netzer, T.~Wang, A.~Coates, A.~Bissacco, B.~Wu, and A.~Y. Ng, ``Reading
  digits in natural images with unsupervised feature learning,'' 2011.

\bibitem{ororbia2020}
A.~Ororbia and D.~Kifer, ``The neural coding framework for learning generative
  models,'' {\em arXiv:2012.03405}, 2020.

\bibitem{telesford2011ubiquity}
Q.~K. Telesford, K.~E. Joyce, S.~Hayasaka, J.~H. Burdette, and P.~J. Laurienti,
  ``The ubiquity of small-world networks,'' {\em Brain Connectivity}, vol.~1,
  no.~5, pp.~367--375, 2011.

\bibitem{Lillicrap20}
T.~Lillicrap, A.~Santoro, L.~Marris, C.~Akerman, and G.~Hinton,
  ``Backpropagation and the brain,'' {\em Nature Reviews Neuroscience},
  vol.~21, 04 2020.

\bibitem{whittington2019theories}
J.~C. Whittington and R.~Bogacz, ``Theories of error back-propagation in the
  brain,'' {\em Trends in Cognitive Sciences}, 2019.

\bibitem{scellier17}
B.~Scellier and Y.~Bengio, ``Equilibrium propagation: Bridging the gap between
  energy-based models and backpropagation,'' {\em Frontiers in Computational
  Neuroscience}, vol.~11, p.~24, 2017.

\bibitem{scellier2018generalization}
B.~Scellier, A.~Goyal, J.~Binas, T.~Mesnard, and Y.~Bengio, ``Generalization of
  equilibrium propagation to vector field dynamics,'' {\em arXiv:1808.04873},
  2018.

\bibitem{rbm}
R.~Salakhutdinov, A.~Mnih, and G.~Hinton, ``Restricted {B}oltzmann machines for
  collaborative filtering,'' in {\em Proceedings of the 24th International
  Conference on Machine Learning}, 2007.

\bibitem{salakhutdinov2009deep}
R.~Salakhutdinov and G.~Hinton, ``Deep {B}oltzmann machines,'' in {\em
  Artificial Intelligence and Statistics}, pp.~448--455, PMLR, 2009.

\bibitem{hinton2009deep}
G.~E. Hinton, ``Deep belief networks,'' {\em Scholarpedia}, vol.~4, no.~5,
  p.~5947, 2009.

\bibitem{Hopfield82}
J.~J. Hopfield, ``Neural networks and physical systems with emergent collective
  computational abilities,'' {\em Proceedings of the National Academy of
  Sciences}, vol.~79, 1982.

\bibitem{Hopfield84}
J.~J. Hopfield, ``Neurons with graded response have collective computational
  properties like those of two-state neurons,'' {\em Proceedings of the
  National Academy of Sciences}, vol.~81, 1984.

\bibitem{auksztulewicz2016repetition}
R.~Auksztulewicz and K.~Friston, ``Repetition suppression and its contextual
  determinants in predictive coding,'' {\em Cortex}, vol.~80, 2016.

\bibitem{lotter2016deep}
W.~Lotter, G.~Kreiman, and D.~Cox, ``Deep predictive coding networks for video
  prediction and unsupervised learning,'' {\em arXiv:1605.08104}, 2016.

\bibitem{watanabe2018illusory}
E.~Watanabe, A.~Kitaoka, K.~Sakamoto, M.~Yasugi, and K.~Tanaka, ``Illusory
  motion reproduced by deep neural networks trained for prediction,'' {\em
  Frontiers in Psychology}, vol.~9, p.~345, 2018.

\bibitem{hohwy2008predictive}
J.~Hohwy, A.~Roepstorff, and K.~Friston, ``Predictive coding explains binocular
  rivalry: An epistemological review,'' {\em Cognition}, vol.~108, no.~3, 2008.

\bibitem{weilnhammer2017predictive}
V.~Weilnhammer, H.~Stuke, G.~Hesselmann, P.~Sterzer, and K.~Schmack, ``A
  predictive coding account of bistable perception-a model-based {fMRI}
  study,'' {\em PLoS Computational Biology}, vol.~13, no.~5, 2017.

\bibitem{feldman2010attention}
H.~Feldman and K.~Friston, ``Attention, uncertainty, and free-energy,'' {\em
  Frontiers in Human Neuroscience}, vol.~4, 2010.

\bibitem{kanai2015cerebral}
R.~Kanai, Y.~Komura, S.~Shipp, and K.~Friston, ``Cerebral hierarchies:
  {P}redictive processing, precision and the pulvinar,'' {\em Philosophical
  Transactions of the Royal Society B: Biological Sciences}, vol.~370, 2015.

\bibitem{millidge2020relaxing}
B.~Millidge, A.~Tschantz, A.~Seth, and C.~L. Buckley, ``Relaxing the
  constraints on predictive coding models,'' {\em arXiv:2010.01047}, 2020.

\bibitem{ororbia2019biologically}
A.~G. Ororbia and A.~Mali, ``Biologically motivated algorithms for propagating
  local target representations,'' in {\em Proc.~AAAI}, vol.~33, pp.~4651--4658,
  2019.

\bibitem{ororbia2022active}
A.~Ororbia and A.~Mali, ``Active predicting coding: Brain-inspired
  reinforcement learning for sparse reward robotic control problems,'' {\em
  arXiv preprint arXiv:2209.09174}, 2022.

\bibitem{strukov2008missing}
D.~B. Strukov, G.~S. Snider, D.~R. Stewart, and R.~S. Williams, ``The missing
  memristor found,'' {\em Nature}, vol.~453, no.~7191, pp.~80--83, 2008.

\bibitem{kendall2020training}
J.~Kendall, R.~Pantone, K.~Manickavasagam, Y.~Bengio, and B.~Scellier,
  ``Training end-to-end analog neural networks with equilibrium propagation,''
  {\em arXiv:2006.01981}, 2020.

\bibitem{sacramento2018dendritic}
J.~Sacramento, R.~P. Costa, Y.~Bengio, and W.~Senn, ``Dendritic cortical
  microcircuits approximate the backpropagation algorithm,'' in {\em Advances
  in Neural Information Processing Systems}, pp.~8721--8732, 2018.

\bibitem{Krotov16}
D.~Krotov and J.~J. Hopfield, ``Dense associative memory for pattern
  recognition,'' in {\em Advances in Neural Information Processing Systems},
  2016.

\bibitem{kendall2020}
J.~Kendall, R.~Pantone, K.~Manickavasagam, Y.~Bengio, and B.~Scellier,
  ``Training end-to-end analog neural networks with equilibrium propagation,''
  {\em arXiv preprint arXiv:2006.01981}, 2020.

\bibitem{wright2022deep}
L.~G. Wright, T.~Onodera, M.~M. Stein, T.~Wang, D.~T. Schachter, Z.~Hu, and
  P.~L. McMahon, ``Deep physical neural networks trained with
  backpropagation,'' {\em Nature}, vol.~601, no.~7894, pp.~549--555, 2022.

\end{thebibliography}
\bibliographystyle{ieeetr}

\section*{Checklist}

\begin{enumerate}

\item For all authors...
\begin{enumerate}
  \item Do the main claims made in the abstract and introduction accurately reflect the paper's contributions and scope?
    \answerYes{}
  \item Did you describe the limitations of your work?
    \answerYes{}
  \item Did you discuss any potential negative societal impacts of your work?
    \answerNA{}
  \item Have you read the ethics review guidelines and ensured that your paper conforms to them?
    \answerYes{}
\end{enumerate}

\item If you are including theoretical results...
\begin{enumerate}
  \item Did you state the full set of assumptions of all theoretical results?
    \answerNA{}
        \item Did you include complete proofs of all theoretical results?
    \answerNA{}
\end{enumerate}

\item If you ran experiments...
\begin{enumerate}
  \item Did you include the code, data, and instructions needed to reproduce the main experimental results (either in the supplemental material or as a URL)?
    \answerNo{The code and the data are proprietary.}
  \item Did you specify all the training details (e.g., data splits, hyperparameters, how they were chosen)?
    \answerYes{}
        \item Did you report error bars (e.g., with respect to the random seed after running experiments multiple times)?
    \answerYes{}
        \item Did you include the total amount of compute and the type of resources used (e.g., type of GPUs, internal cluster, or cloud provider)?
    \answerYes{See Supplementary material}
\end{enumerate}

\item If you are using existing assets (e.g., code, data, models) or curating/releasing new assets...
\begin{enumerate}
  \item If your work uses existing assets, did you cite the creators?
    \answerYes{}
  \item Did you mention the license of the assets?
    \answerNA{}{}
  \item Did you include any new assets either in the supplemental material or as a URL?
    \answerNA{}{}
  \item Did you discuss whether and how consent was obtained from people whose data you're using/curating?
    \answerNA{}{}
  \item Did you discuss whether the data you are using/curating contains personally identifiable information or offensive content?
    \answerNA{}{}
\end{enumerate}

\item If you used crowdsourcing or conducted research with human subjects...
\begin{enumerate}
  \item Did you include the full text of instructions given to participants and screenshots, if applicable?
    \answerNA{}{}
  \item Did you describe any potential participant risks, with links to Institutional Review Board (IRB) approvals, if applicable?
    \answerNA{}{}
  \item Did you include the estimated hourly wage paid to participants and the total amount spent on participant compensation?
    \answerNA{}{}
\end{enumerate}

\end{enumerate}

\newpage

\appendix

\begin{algorithm}[t]
\footnotesize
    \caption{Learning the external stimulus $\bar s$} \label{algo:IL}
    \begin{algorithmic}[1]
    \REQUIRE  $(x_{1,t},\dots,x_{d,t})$ is fixed to $(s_{1},\dots,s_{d})$.
    \FOR{$t=0$ to $T$}
        \FOR{each vertex $i$}
            \STATE update $x_{i,t}$ to  minimize $\mathcal{E}_{t}$ via Eq.~\eqref{eq:x_update}
        \ENDFOR
        \IF{$t= T$} 
            \STATE update every $\theta_{i,j}$ to minimize $\mathcal{E}_{t}$ via Eq.~\eqref{eq:theta_update}.
        \ENDIF
    \ENDFOR;
    \end{algorithmic}
\end{algorithm}

\section{A Discussion on Biological Plausibility}

\begin{figure*}[t]
\medskip 
    \centering
	\includegraphics[width=1.0\textwidth]{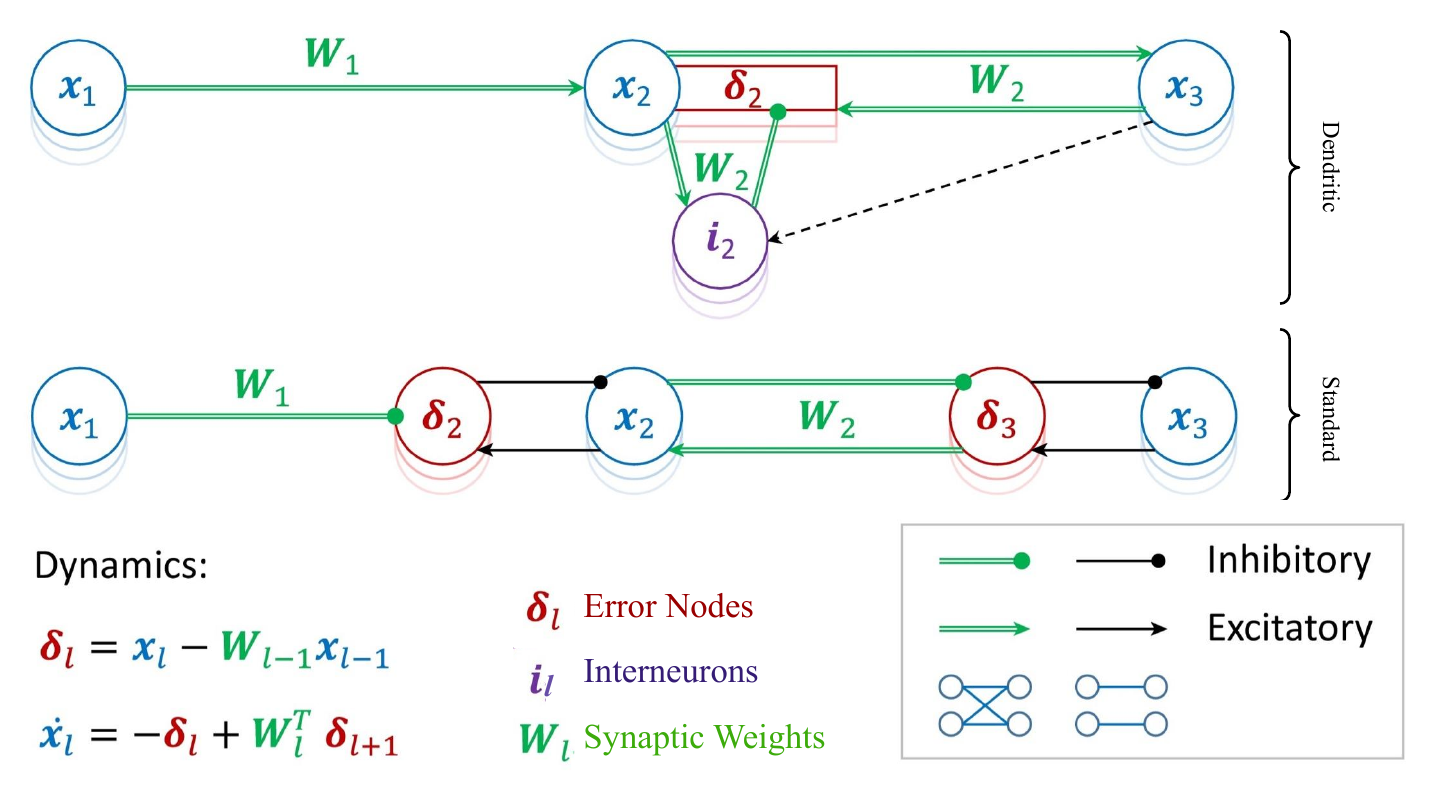}
\caption[short]{Standard and dendritic neural implementation of predictive coding. The dendritic implementation makes use of interneurons $i_l = W_l x_l$ (according to the notation used in the figure). Both implementations have the same equations for all the updates, and are hence equivalent; however, dendrites allow a neural implementation that does not take error nodes into account, improving the biological plausibility of the model. Figure taken and adapted from \cite{whittington2019theories}.
}
\label{fig:den}
\end{figure*}

In the literature, there is often a disagreement on when a specific algorithm can be considered biologically plausible. This follows, as every computer simulation fails to be completely equivalent to every aspect on how the brain works, as there will always be some details that make the simulation implausible. Hence, it is normally assumed that an algorithm is biologically plausible when it satisfies a list of properties that are also satisfied in the brain. Different works consider different properties. In our case, we consider as list of minimal properties that a learning rule should satisfy, the ones that allow to have a possible neural implementation, such as local computations and lack of a global control signal to trigger the operations. However, the neural implementation proposed in Fig.~\ref{fig:fully} takes error nodes into account, often considered implausible from the biological perspective \cite{sacramento2018dendritic}. Even so, the biological plausibility of our model is not affected by this: it is in fact possible to map PCNs on a different neural architecture, in which errors are encoded in apical dendrites rather than separate neurons \cite{sacramento2018dendritic,whittington2019theories}. Graphical representations of the differences between the two implementations is given in Fig.~\ref{fig:den}, taken (and adapted) from \cite{whittington2019theories}.

\section{Methodology and Further Experiments}

Compared to backpropagation (BP), predictive coding (PC) allows for more flexibility in the definition, training, and evaluation of the model. The experiments reported in this paper show the best results achieved on each specific task and, as a consequence, only the effects of a specific set of hyperparameters. Therefore, the complete range of possibilities that exist in PC has not been displayed, however, those alternative configurations may be helpful in other scenarios. A pseudocode that describes the learning process of PC graphs is given in Algorithm~\ref{algo:IL}.

\subsection{Architectures and Hyperparameters}
In this section, we provide a detailed description of the models and parameters used to obtain the results in the various generation tasks presented in this work, to guarantee their reproducibility. Note that our goal was to compare the performance of different models, hence we compare networks that have a similar number of parameters. We now briefly summarize the PC graphs used in this work:

\begin{itemize}
    \item \textbf{Fully connected networks:} The experiments in the paper body are obtained by using a fully connected graph with $2000$ vertices, trained with $794$ sensory vertices for classification and generation tasks ($784$ pixels plus  a 1-hot vector for the $10$ labels), and $784$ sensory vertices for reconstruction and denoising. For colored images, we used a network with $5000$ vertices. We  trained every model for $20$ epochs, and reported the best results using early stopping. As learning rates, we  used $\alpha \,{\in}\, \{1,0.5\}$ for the value nodes, and $\eta \in \{0.0001,$ $0.00005\}$ for the weights, and a weight decay $\lambda = \{0.01,0.001,0.0001,0\}$. To conclude, we computed each query using $T \,{=}\, 2000$, making sure that the energy had converged before reaching that~value.

    \item \textbf{Feedforward network:} A network composed by a sequence of $L$ fully connected layers of dimension $H$. The best results were achieved with $L \in \{3, 4\}$ and $H=512$ for MNIST and $H=1024$ on Fashion\-MNIST. We did not experience any benefits in adding extra layers, as it only resulted in higher convergence times. The width, instead, directly determines the quality of the images produced: as expected, very narrow networks fail to store enough information to accurately reconstruct (or denoise) the input images. However, wide networks manifest sub-optimal performance as well. This follows, as having more parameters allows the network to easily overfit. As a consequence, the generation process is less stable, and the images can appear noisier and composed by strokes belonging to different classes. Using a strong weight decay alleviates these problems, as we will later discuss.
    
    \item \textbf{Recurrent network:} A recurrent layer consists of a layer whose output is transformed by a non-linear transformation and fed in input to the layer. The recurrent networks used in this paper consist of two recurrent layers (for a total of four non-linear transformations) with hidden dimension $H=512$ when trained on MNIST, and $H=1024$ when trained on FashionMNIST. The behaviour, given the choice of width and depth, seems similar to feedforward networks. The performance, however, seems to be less impacted by the usage of wide layers. This is due to the recurrent connections that establish more constraints, and thus stability.

    \item \textbf{Assembly of neurons:} As stated in the paper body, we  used models with $4$ clusters with $3000$ vertices each, connected in a feedforward way. As sparsity and top-k constants, we  used $p=0.1$ and $k=0.2$, and performed the same generative experiments. Again, we  trained each model for $20$ epochs, and reported the best results using early stopping. As learning rates, we  used $\alpha \in \{1,0.5\}$ for the value nodes, and $\eta \in \{0.0001,$ $0.00005\}$ for the weights. To conclude, we  computed each query using $T = 2000$, making sure that the energy had converged before reaching that~value.
    
    \item \textbf{Autoencoders:} The autoencoder was defined using the same shape as the feedforward networks: it is as a fully connected network with $L \in \{3, 4\}$ hidden layers of width ${H \in \{256,512,1024\}}$. In this way, the structure and the number of parameters directly correspond to the feedforward network trained using predictive coding. It was trained through BP using the \textit{Adam} optimizer, with learning rate $\alpha = 1e^{-4}$ and weight decay of parameter ${\lambda \in \{1e^{-2}, 1e^{-4}, 1e^{-6}, 0\}}$ (the best results were achieved with the lowest value).
\end{itemize}

As predictive coding requires two sets of updatable parameters, the value nodes $x_{i,t}$ and the weights $\theta_{i,j}$, we defined two separate optimizers. The learning rate for the weights was set to $\alpha = 1e^{-4}$, and the optimizer algorithm chosen was \textit{Adam} (as for the autoencoder). We experimented with different values of weight decays, noticing how the final performance is highly affected by this value. For the given tasks, the best results were achieved with \textit{weight decay} $= 1e^{-2}$. Instead, the learning rate for the value nodes  was set to $\gamma = 1.0$, and optimized using \textit{SGD}. 
To conclude, we have tested different activation functions; the most promising seems to be \textit{HardTanh}.


\subsection{Feedforward vs.\ Recursive Networks}

\begin{figure}
    \centering
    \includegraphics[width=\linewidth]{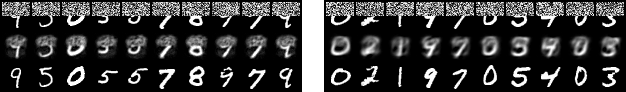}
    \vspace*{-2ex}
    \caption{Reconstruction using query by conditioning on the whole output layer. The performance of feedforward networks (left) is noticeably improved by using recurrent connections (right), as the reconstructed images do not overfit the noise, but resemble plausible, albeit noisy, digits.}
    \label{fig:rec_half_noise}
\end{figure}

\begin{figure}
    \centering
    \includegraphics[width=\linewidth]{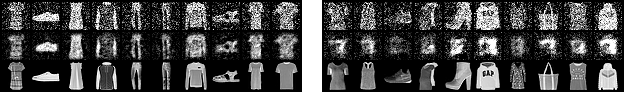}
    \vspace*{-2ex}
    \caption{Reconstruction using query by conditioning using FashionMNIST samples after training on MNIST. Feedforward networks (left) simply overfit (i.e., reproduce without performing any modification) the input samples, despite being unrelated to the training data. Recurrent networks, instead, reproduce an unrecognisable and shady image, showing that they do not recognize the input samples, as they are not stable data points.}
    \label{fig:mnist_denoise_fashion}
\end{figure}

In this work, we highlighted how in different situations, one may prefer to query by conditioning or by initialization. As a rule of thumb, conditioning means that we expect the partial data given to the network to be correct and be recognized as a \textit{memory}, by being reconstructed by the network without modifications. Therefore, it makes sense to use it in the reconstruction generative task. Instead, when performing image denoising, we do not want the network to recall the noisy image from its memory, instead, we are asking it to retrieve the memory (or to generate a realistic sample), representing a plausible image, that is the closest to the noisy input. It makes therefore sense to only initialize the output layer, giving the network a direction to follow and let it evolve unconstrained. However, it may not always be clear which querying technique is most preferable. 
A~desirable behavior may be using the network to identify which querying data are realistic (i.e., similar to the training samples) and which not. Ideally, we would like the network to perfectly fit previously seen data points, while struggling to reconstruct unfamiliar shapes. We tested both the feedforward and recursive networks by training them on the MNIST dataset and querying them by conditioning the output layer with a full-size image composed by half uniform noise and half digit. The results are reported in Fig.~\ref{fig:rec_half_noise}. We can see how feedforward networks easily fit the noise, reconstructing the two halves independently. On the other hand, employing recurrent connections (and thus imposing stricter constraints) forces the network to reconstruct the  image as a whole. We can see a similar behavior in Fig.~\ref{fig:mnist_denoise_fashion}, where networks trained on MNIST are use to denoise FashionMNIST images. Feedforward networks easily overfit the input samples. Recurrent networks, instead, correctly do not recognize the given images and reconstruct an unrelated and confused blob. In this last case, it would therefore be possible to distinguish between familiar and unfamiliar images by computing the distance between the input and~output~images.

\subsection{Importance of Weight Decay}
As previously mentioned, weight decay plays a fundamental role in determining the properties of the reconstructed images. Compared to other tasks (e.g., classification) or models (e.g., autoencoder trained by BP), a higher value of weight decay seems to be necessary when training with PC. From our experiments, weight decay prevents the networks from overlearning the task that they are trained on (i.e., reproduce any image that they are given in input), and instead allows them to ``understand'' the several concept classes of each dataset. This behaviour makes it possible to generalize their knowledge to new and unseen tasks, such as the denoising and reconstructing tasks seen in this work. It is worth noticing how, when optimizing for a single specific problem (e.g., image recognition), lower values of weight decay seem to be more effective.

\begin{figure}
    \centering
    \includegraphics[width=0.95\linewidth]{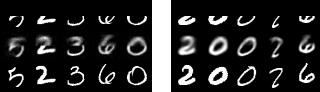}
    \caption{Reconstructed images given the label and by conditioning the bottom half. Using low weight decay values (left) causes the two halves of the images to be uncorrelated. As a result, each digit is composed by almost unrelated lines. Contrarily, with higher values (right), each image is correctly generated. }
    \label{fig:low_vs_high_wd}
\end{figure}

To show this, we trained a recurrent network to reconstruct images by conditioning the bottom half of the output layer and giving the target class label in input. The result is that, with low weight decay, the network treats each half of the image independently, reconstructing the bottom part by fitting the conditioning data and the top half using the given label. It can be observed that there is no relation between the two halves. With higher weight decay, instead, we can see that the image is reconstructed as a whole, incorporating both the information provided via the label and the conditioning data~(Fig.~\ref{fig:low_vs_high_wd}).

\section{Associative Memory Experiments}

\begin{figure*}[t]
\medskip 
    \centering
	\includegraphics[width=1.0\textwidth]{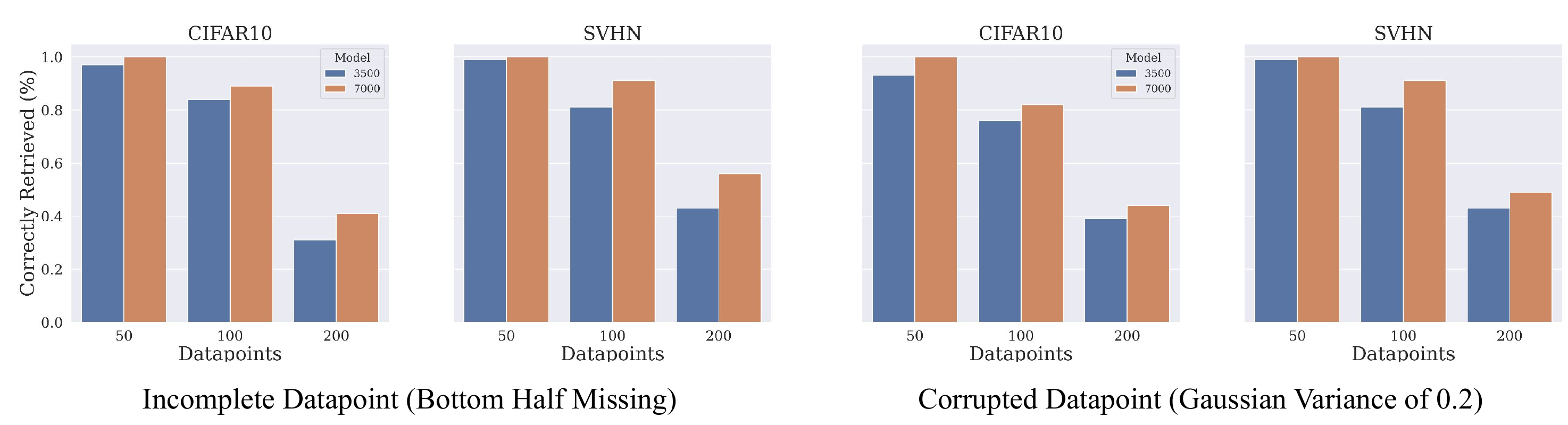}
	\vspace*{-3ex}
\caption[short]{Number of correctly retrieved data points when presented with incomplete or corrupted variations. We  used datasets of $\{50,100,200\}$ images of the CIFAR10 and SVHN datasets, and trained on fully connected PC graphs of size $\{3500,7000\}$ vertices.
}
\label{fig:memo}
\end{figure*}

In the paper body, we claimed that a fully connected PC graph is able to perform associative memory (AM) experiments. To show this, we  trained fully connected PC graphs with $\{3500,7000\}$ vertices on different subsets of cardinality $\{50,100,200\}$ of CIFAR10 and FashionMNIST. Then, we  used query by initialization and conditioning to retrieve the original memories. In this setting, we  considered a memory to be retrieved if the mean squared error between the original training point and its reconstruction is less than $0.001$. As corruption, we  either removed the top half of the image, or corrupted it with Gaussian noise of mean zero and variance $0.2$. The results are shown in Fig.~\ref{fig:memo}.

\textbf{Results:} The experiments show that our model is able to well store and retrieve memories, even when tested on colored images. The reconstruction quality, as expected, decreases when adding more memories, and improves when adding more parameters to the model. As hyperparameters, we used $\eta = 0.0001$, $\alpha = 0.5$, and $T=5$.

\section{Classification Results}

\begin{table*}[t]
    \centering
	\begin{tabular}{lccccc}
		\toprule
		 Model &  PC & RBM & DAM  & BP   \\
		\toprule
		MNIST    & 98.47 $\pm$ 0.12  & 94.12 $\pm$ 0.59 & 98.58  & 98.41 $\pm$ 0.18   \\
		FashionMNIST    & 89.92 $\pm$ 0.23  & 86.98 $\pm$ 0.49 &  90.22 ± 0.27  & 90.29 $\pm$ 0.33 \\
		SVHN & 88.99 $\pm$ 0.26 & 85.09 $\pm$ 0.87 & 86.77 $\pm$ 0.22  & 89.31 $\pm$ 0.09 \\
		CIFAR10 & 56.23 $\pm$ 3.36 & 41.12 $\pm$ 3.88 & 46.06 $\pm$ 2.77 & 59.11 $\pm$ 2.47 \\
		\bottomrule
	\end{tabular}
	\caption{Test accuracy of multilayer PCNs (i.e., feedforward PC graphs)  on MNIST, FasionMNIST, SVHN, and CIFAR10. The results are compared against popular models in the literature:  restricted Boltzmann machines (RBMs) \cite{rbm}, dense associative memories (DAMs) \cite{Krotov16}, and multilayer perceptrons (MLPs) trained with BP \cite{rumelhart1986learning}. Classification on MNIST using DAM does not report variance, as it is taken from the original work, and the authors only report the average.}
	\label{tab:acc_mlp}
	\end{table*}

In the paper body, we  stated that multilayer PCNs are known to perform similarly to BP on classification. Here, we  tested this, and compared against popular models in the literature, such as restricted Boltzmann machines (RBMs) \cite{rbm} and \emph{dense associative memories} (DAMs) \cite{Krotov16}. Overall, PCNs are the only models able to perform similarly to BP on the test set. We  performed experiments on $4$ datasets: MNIST, FashionMNIST, SVHN, and CIFAR10, and the results are  in Table~\ref{tab:acc_mlp}.

\textbf{Setup:} The networks trained using PC and BP have $L=\{2,3\}$ and 256 hidden neurons each. They are trained using Adam optimization, a weight decay $\lambda \in \{0.001,0.0001,0\}$, and the learning rate for the weights $\alpha \in \{0.001,0.0001\}$. We report the best average results in  Table~\ref{tab:acc_mlp}. For the RBM, we  used a model with $512$ hidden nodes, and for the DAM, we copied the official implementation provided by the authors, with the same hyperparameters.

\section{Restricted Boltzmann Machines}

\begin{figure}
    \centering
    \includegraphics[width=\linewidth]{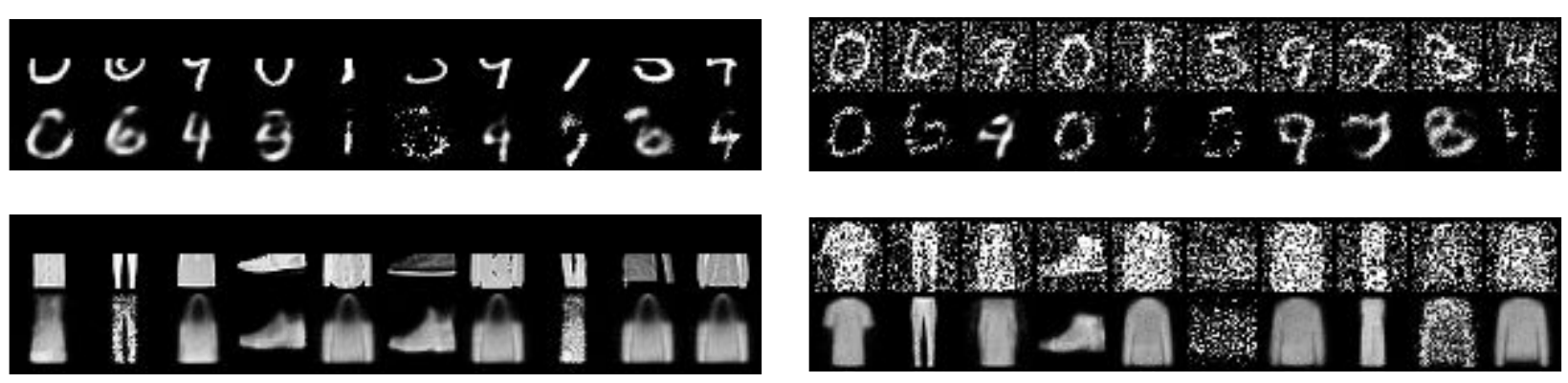}
    \vspace*{-3ex}
    \caption{Reconstructed and denoised images using RBMs. }
    \label{fig:rbm}
\end{figure}

To provide a full comparison between the generation capabilities of our model and existing ones in the literature, we trained a different RBM, and performed both reconstructions and denoising tasks. The results are in Fig.~\ref{fig:rbm}. Particularly, they show that RBMs sometimes fail to retrieve the correct image, returning a blurry cloud of points in denoising, and tend to often return the same image even when presented with different inputs in reconstruction ones. This problem was consistent in different batches and parametrizations of RBMs, and never happened in any of the models that we have proposed.

\textbf{Setup:} We trained several RBMs with $h \,{\in}\, \{256,$ $512,$ $1024\}$ hidden nodes, and performed $\{1,2,5,10\}$ \emph{Gibbs samplings}. We always picked the best result.

\section{High Levels of Noise}

\begin{figure*}[t]
\medskip 
    \centering
	\includegraphics[width=1.0\textwidth]{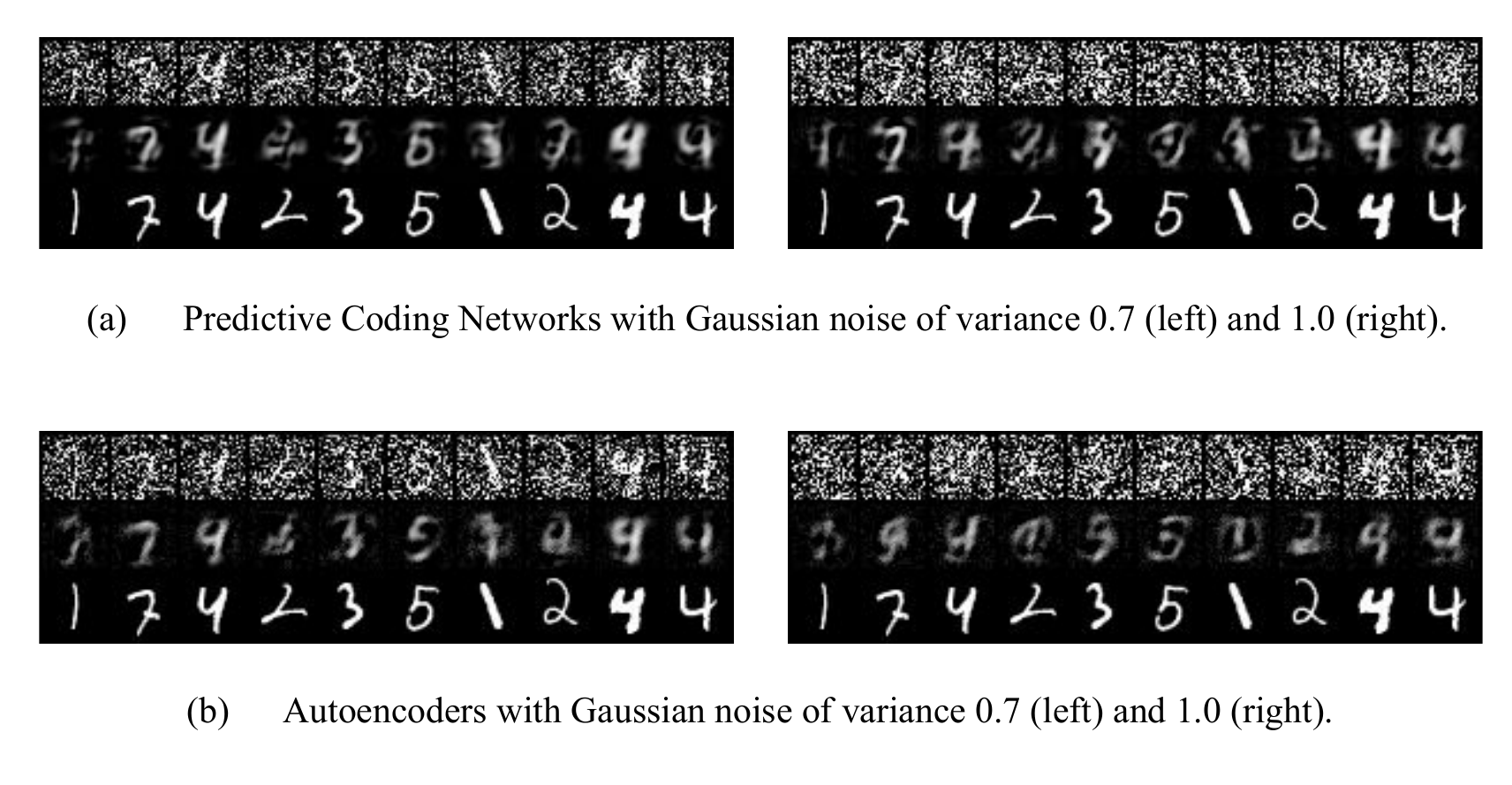}
\caption[short]{Denoising tasks when presented with high levels of noise.}
\label{fig:high_noise}
\end{figure*}

Here, we push the limits of the model in denoising tasks, where the variance of the Gaussian noise is high enough such that it is often hard for a human evaluator to distinguish different numbers. Particularly, we use a $3$ layer PCN with $256$ hidden neurons, and we test it against an autoencoder with the same parametrization. The results, provided in Fig.~\ref{fig:high_noise} show that both models fail to reconstruct some examples, and the reconstructed ones are noisy. However, we note that PCNs are able to distinguish more numbers than autoencoders, and hence have a better overall performance in this task.

\section{Efficiency of the Model}

Training a deep PC network is almost as fast as training deep neural networks with backpropagation. This is despite the fact that every hardware and library is highly optimized for the latter. However, while not faster today, efficiency is an interesting property of PC graphs, and many other neuroscience-inspired learning methods, such as equilibrium and target propagation: all these algorithms are slower than backpropagation; however, they are extremely promising with respect to future developments on the hardware side. In fact, they would allow to train deep neural networks in an end-to-end fashion on physical chips, such as analog circuits \cite{kendall2020}. This is something that is not possible to do with backpropagation: in \cite{wright2022deep}, the authors implement exact backpropagation on physical chips. However, the process is quite slow, as there is the need of a digital control signal at every layer of the network. This is due to the sequential structure of deep models, where every operation of a layer has to (1) \emph{wait} for the information of all the previous (following during the backward pass) layers, and (2) be saved in memory via a von-Neumann digital device. The situation would be completely different if using methods that would allow to train neural networks end-to-end, i.e., without any digital component, on the same chip: in this case, the learning process would be much faster, and would not need any external control to be performed. This is possible by using PC. However, despite potential applications on physical chips, PC is also fast on current GPUs, and hence this is not an obstacle towards applications. We now show multiple plots that shot the training and inference times of multiple PC models. Note that these results are obtained by using an implementation that does not make use of the full parallelization capabilities of PC, as this is not supported by standard deep learning frameworks (in our case, Pytorch). Hence, the proposed plots largely overestimate the actual efficiency of PCNs that can be obtained via a correct implementation.

\begin{figure}[t]
\medskip 
    \centering
	\includegraphics[width=0.95\textwidth]{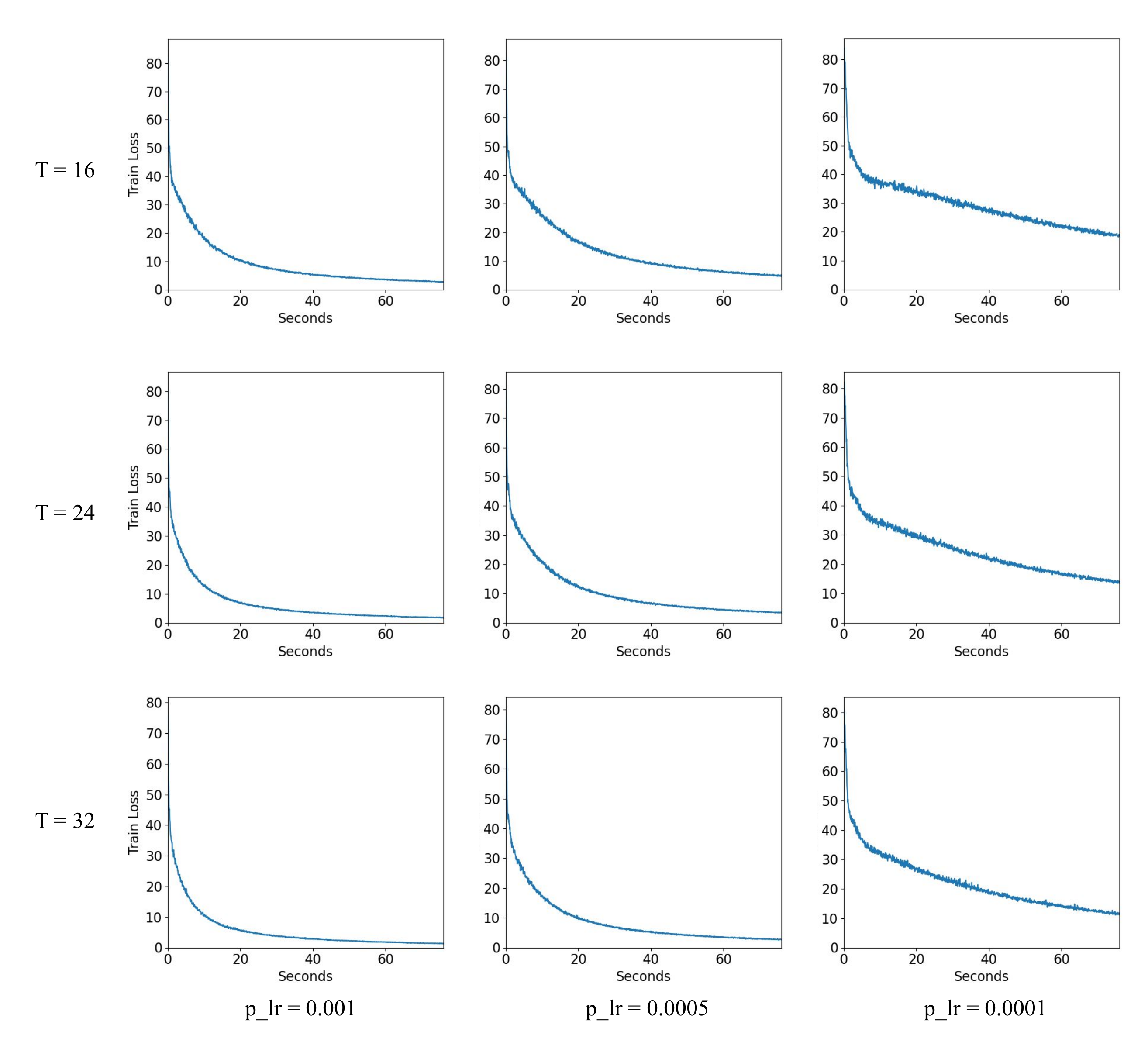}
\caption[short]{Energy as a function of time (in seconds $s$) for different hyperparameters during training.
}
\label{fig:train_speed}

\medskip \medskip\medskip
    \centering
	\includegraphics[width=0.8\textwidth]{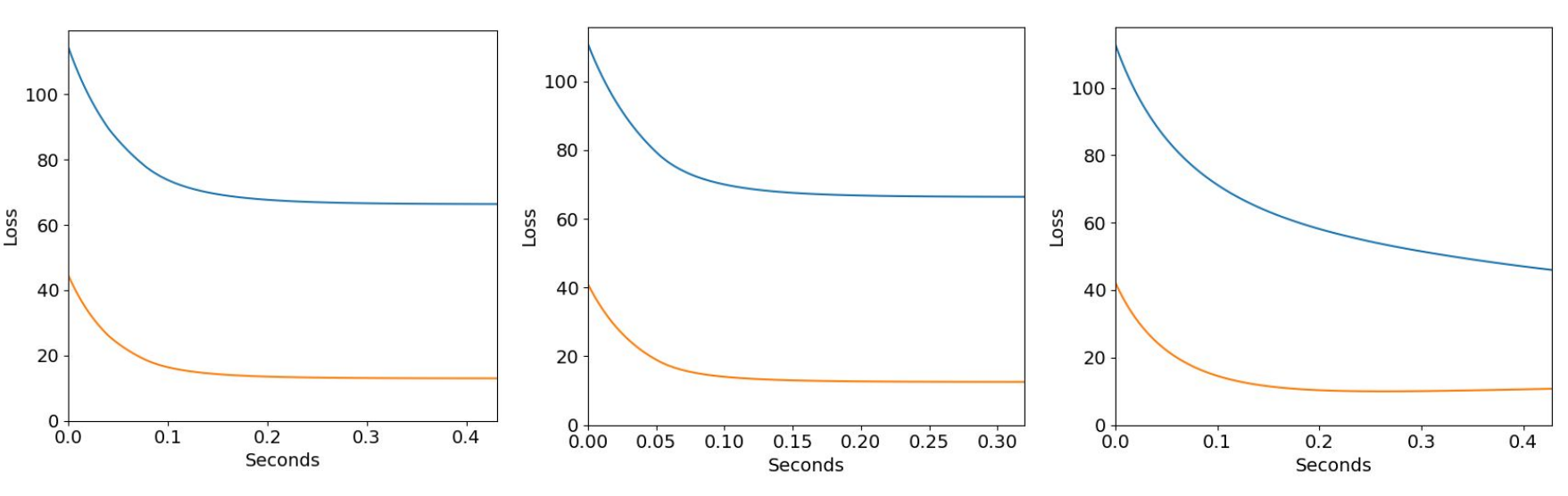}
\caption[short]{Total energy (blue) and loss (orange) of retrieval (left) and denoising (centre) tasks on a 3-generative model with $512$ hidden neurons per layer. On the right, retrieval of the same model, with added recurrent connections.}
\label{fig:test_speed}
\vspace*{-1ex}
\end{figure}

\begin{figure}[t]
\medskip 
    \centering
	\includegraphics[width=1\textwidth]{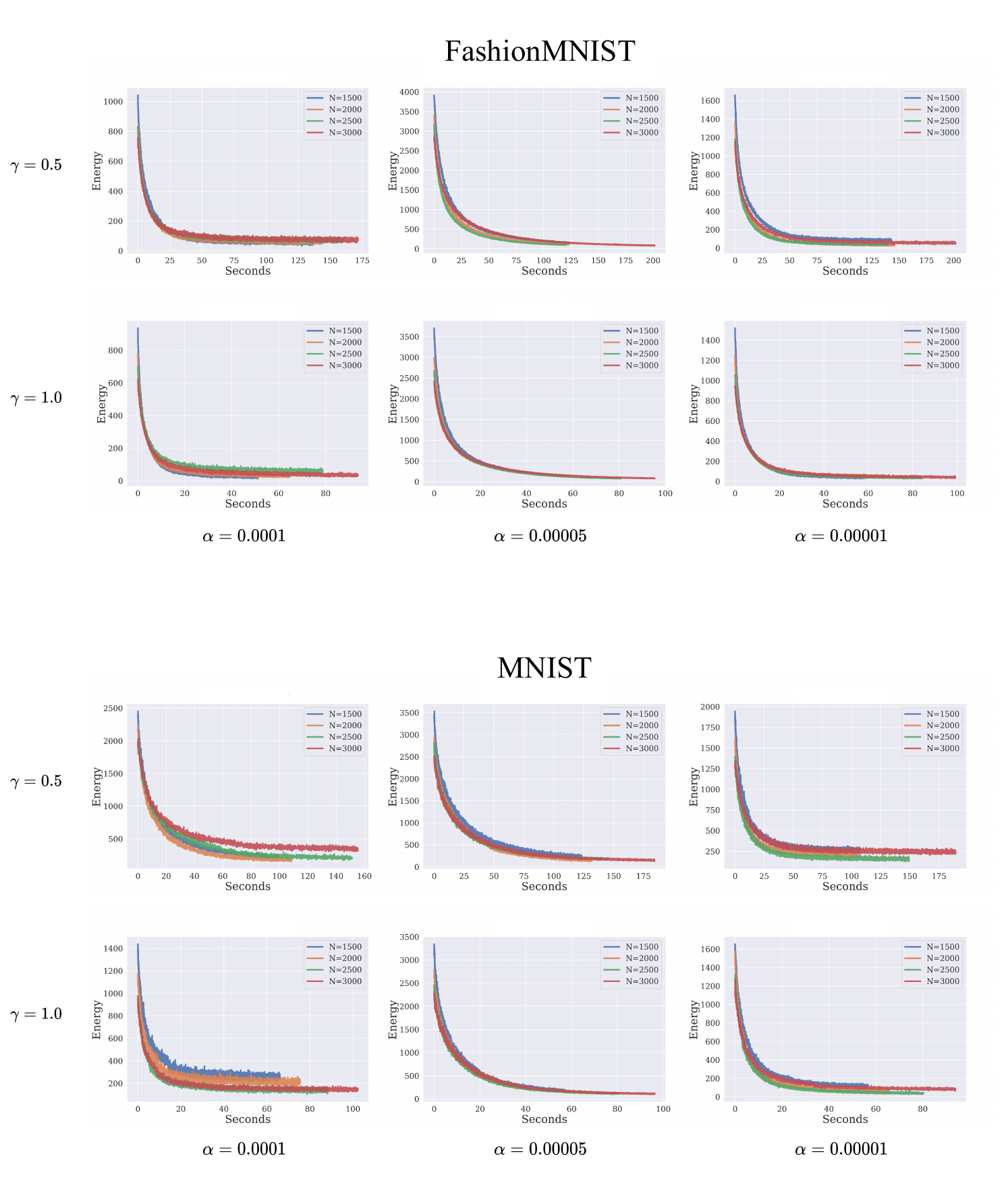}
\caption[short]{Energy as a function of time (in seconds) for different hyperparameters during training of a PC graph.
}
\label{fig:train_graph}
\end{figure}

\paragraph{Experiments:} Here, we provide multiple plots that show that PC graphs quickly converge to a stationary point. Particularly, we show that the provided experiments are fast: training a recurrent 3-layer PCN takes about 1 minute on an RTX Titan, as shown in the plots in Fig.~\ref{fig:train_speed}. Same for testing: reconstructing/denoising an image takes 0.1/0.3 secs, as shown by the plots provided in Fig.~\ref{fig:test_speed}. Hence, the proposed models are robust to hyperparameter changes and converge rapidly. All the proposed plots are generated via training and testing on a multilayer generative PCN with $3$ layers and $512$ hidden neurons per layer. We also provide the convergence plot of $48$ different PC graphs, of different parametrizations ($N \in \{1500,2000,2500,3000\}$), learning rates ($\alpha \in \{0.0001, 0.00005, 0.00001\}$) and integration steps ($\gamma \in \{1.0, 0.5\}$), on both MNIST and FashionMNIST. As shown in Fig.~\ref{fig:train_graph}, PC graphs always and quickly converge.

\end{document}